%% file: main.tex
\documentclass[english]{article}
\usepackage{kr}

\usepackage{times}
\usepackage{soul}
\usepackage{url}
\usepackage[hidelinks]{hyperref}
\usepackage[utf8]{inputenc}
\usepackage[small]{caption}
\usepackage{graphicx}
\usepackage{amsmath}
\usepackage{amsthm}
\usepackage{amssymb}
\usepackage{booktabs}
\usepackage{enumitem}
\usepackage{algorithm}
\usepackage{algorithmic}
\usepackage{babel}
\usepackage{todonotes}
\usepackage{xspace}
\usepackage{stmaryrd}
\usepackage{mathtools}
\usepackage{tabularx}
\usepackage{pgfplots}
\usepackage{subcaption}
\usepackage{csquotes}

\pgfplotsset{compat=1.3}

\newtheorem{example}{Example}
\newtheorem{theorem}{Theorem}
\newtheorem{lemma}{Lemma}
\newtheorem{proposition}{Proposition}
\newtheorem{definition}{Definition}
\newenvironment{sketch}{%
  \proof}{\endproof}

\newcommand{\tagtext}[2]{%
  \expandafter\def\csname tagtext@#1\endcsname{#2}%
  \hypertarget{#1}{#2}%
}

\newcommand{\linktext}[1]{%
  \hyperlink{#1}{\csname tagtext@#1\endcsname}%
}
\input{macros}

\title{Automated Planning with Ontologies under Coherence Update Semantics}

\author{%
Stefan Borgwardt$^1$\and
Duy Nhu$^{1}$\and
Gabriele Röger$^2$\\
\affiliations
$^1$Institute of Theoretical Computer Science, TU Dresden, Germany\\
$^2$Department of Mathematics and Computer Science, University of Basel, Switzerland\\
\emails
\{stefan.borgwardt, hoang\_duy.nhu\}@tu-dresden.de,
\{firstname.lastname\}@unibas.ch
}

\begin{document}

\maketitle

\input{parts/abstract}

\input{parts/intro}

\input{parts/pre}

\input{parts/coherence_planning}

\input{parts/exp}
\input{parts/concl}

%% The file kr.bst is a bibliography style file for BibTeX 0.99c
\bibliographystyle{kr}
\bibliography{references}

\clearpage

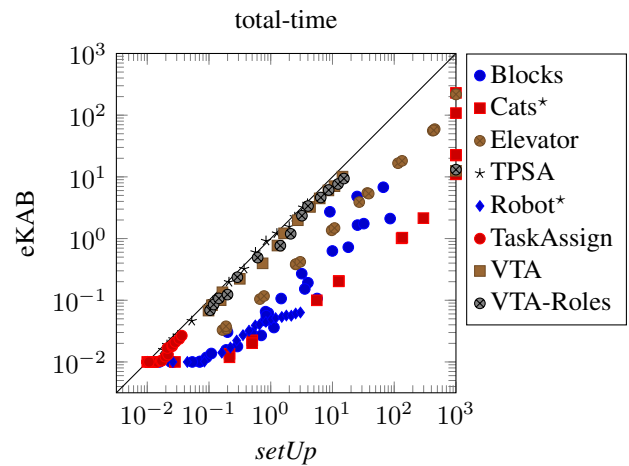
\end{document}

\maketitle
\input{parts/appendix}

% \bibliographystyle{kr}
% \bibliography{references}

%% file: macros.tex
\newcommand{\ttodo}[4]{\ifthenelse{\equal{#1}{inline}}{\todo[inline, author=#2, color = 
#3]{#4}}{\todo[color=#3]{#2: #4}}}

\newcommand{\inlinecite}[1]{\citeauthor{#1} (\citeyear{#1})}

\newcommand{\wlg}{w.l.o.g.\ }

\newcommand{\wrt}{w.r.t.\ }

\newcommand{\ie}{i.e.\ }
\newcommand{\eg}{e.g.\ }

\newcommand{\NC}{\ensuremath{\mathbf{C}}\xspace}
\newcommand{\NR}{\ensuremath{\mathbf{R}}\xspace}
\newcommand{\NI}{\ensuremath{\mathbf{I}}\xspace}
\newcommand{\mA}{\ensuremath{\mathcal{A}}\xspace}
\newcommand{\mE}{\ensuremath{\mathcal{E}}\xspace}
\newcommand{\mF}{\ensuremath{\mathcal{F}}\xspace}
\newcommand{\mG}{\ensuremath{\mathcal{G}}\xspace}
\newcommand{\mH}{\ensuremath{\mathcal{H}}\xspace}
\newcommand{\mI}{\ensuremath{\mathcal{I}}\xspace}
\newcommand{\mK}{\ensuremath{\mathcal{K}}\xspace}
\newcommand{\mO}{\ensuremath{\mathcal{O}}\xspace}
\newcommand{\mP}{\ensuremath{\mathcal{P}}\xspace}
\newcommand{\mR}{\ensuremath{\mathcal{R}}\xspace}
\newcommand{\mT}{\ensuremath{\mathcal{T}}\xspace}
\newcommand{\mU}{\ensuremath{\mathcal{U}}\xspace}
\newcommand{\DLLite}{\ensuremath{\textsl{DL-Lite}}\xspace}

\newcommand{\cc}[1]{\text{\upshape{\textsc{#1}}}\xspace}

\newcommand{\NP}{\cc{NP}}

\newcommand{\ExpTime}{\cc{Exp\-Time}}

\newcommand{\ExpSpace}{\cc{Exp\-Space}}

\DeclarePairedDelimiter{\norm}{\lVert}{\rVert}

\newcommand{\equivto}[2]{\underset{\textstyle\overset{\mkern4mu\rotatebox{90}{$\equiv$}}{#2}}{#1}}

\newcommand{\ex}[1]{\ensuremath{\mathsf{#1}}\xspace}
\newcommand{\funct}{\ensuremath{\texttt{\upshape funct}\,}\xspace}

\newcommand{\numtasks}[1]{\small{(#1)}}
\newcommand{\astar}{\ensuremath{\textrm{A}\!^*}\xspace}
\newcommand{\ffapprox}{\smash{$\widetilde{\text{FF}}$}}
\newcommand{\varzero}{\emph{deriveUp}}
\newcommand{\vartwo}{\emph{setUp}}

%% file: parts/abstract.tex
\begin{abstract}
  Standard automated planning employs first-order formulas under closed-world semantics to achieve a goal with a given set of actions from an initial state.
  We follow a line of research that aims to incorporate background knowledge into automated planning problems, for example by means of ontologies, which are usually interpreted under open-world semantics.
  We present a new approach for planning with \DLLite ontologies that combines the advantages of ontology-based action conditions provided by \emph{explicit-input knowledge and action bases (eKABs)} and ontology-aware action effects under the \emph{coherence update semantics}.
  We show that the complexity of the resulting formalism is not higher than that of previous approaches, and provide an implementation via a polynomial compilation into classical planning.
  An evaluation on existing and new benchmarks examines the performance of a planning system on different variants of our compilation.
\end{abstract}

%% file: parts/intro.tex
\section{Introduction}
Automated planning is a core area within Artificial Intelligence that describes the development of a system through the application of actions \cite{DBLP:books/daglib/0014222}.  
A planning task is defined by an initial state, a set of actions with preconditions and effects on the current state, and a goal condition.
States can be seen as finite first-order (FO) interpretations, and all conditions are specified by FO-formulas that are interpreted on the current state under closed-world semantics.
The objective is to select a sequence of applicable actions to reach the goal.

In the literature, one can find many approaches to add expressive reasoning facilities to planning formalisms \cite{DBLP:conf/aaai/AhmetajCOS14,DBLP:conf/pmai/GiacomoFS24,DBLP:conf/icaps/CorreaGHR24,DBLP:conf/ecai/JohnK24}.
One approach is to use additional logical theories under open-world semantics to describe the possible interactions between objects of a domain of interest.
Here, we are interested in \emph{Description Logics} (\emph{DLs}) and their application in reasoning about the individual states of a system.
The main challenge is to reconcile the open-world nature of DLs and the closed-world semantics employed in classical planning.

\emph{Explicit-input Knowledge and Action Bases} (\emph{eKABs}) combine planning with the description logic \DLLite \cite{DBLP:conf/ijcai/CalvaneseMPS16}.
There, states (\emph{ABoxes}) are interpreted using open-world semantics \wrt a \emph{background ontology} (\emph{TBox}) specifying intensional knowledge using \DLLite axioms.
The background ontology describes constraints on the state, and entails additional facts that hold implicitly.
Such a planning problem can be compiled into the classical \emph{planning domain definition language (PDDL)} using query rewriting techniques \cite{DBLP:conf/ijcai/CalvaneseMPS16}.

\begin{example}\label{ex:blocks}
Consider the following axioms and facts in a blocks world:
\begin{align*}
    % \mT = \{
    &
        \ex{on\_block} \sqsubseteq \ex{on}, \
        \exists \ex{on\_block}^- \sqsubseteq \ex{Block}, \
        \funct \ex{on\_block}, \\
    &
        \ex{on\_table} \sqsubseteq \ex{on}, \
        \exists \ex{on\_table}^- \sqsubseteq \ex{Table}, \
        \ex{Block} \sqsubseteq \lnot \ex{Table}, \\
    &
        \ex{Block} \equiv \exists \ex{on}, \
        \exists \ex{on\_block}^- \sqsubseteq \ex{Blocked}, \\
    &
        \exists\ex{on\_block} \sqsubseteq \lnot\exists\ex{on\_table}, \
    % &
    %     \ex{Holding} \sqsubseteq \ex{Block}, \
    %     \ex{Holding} \sqsubseteq \ex{Blocked}, \
    % &
    %     \ex{Block} \equiv \exists \ex{has\_color}, \
    %     \funct \ex{has\_color}, \
    %     \exists \ex{has\_color}^- \sqsubseteq \ex{Color}
    % \}
% \end{align*}
% \begin{align*}
    % \mA = \{
    % &
        \ex{on\_block}(b_1,b_2), \
        % \ex{on\_table}(b_2,t), \
        \ex{on\_table}(b_3,t)
    % &
    %     \ex{has\_color}(b_1,g), \
    %     \ex{has\_color}(b_2,r)
    % \}
\end{align*}

% Action \ex{paint}: \\ $\lnot\ex{Covered}(b)\land\ex{Color}(c)\land\lnot\ex{has\_color}(b,c)$ \\ $\leadsto$ $\ex{has\_color}(b,c)$ \\ implicit effect: remove previous $\ex{has\_color}(b,c')$

% Action \ex{remove}: \\ $\ex{Block}(b)\land\lnot\ex{Covered}(b)$ \\ $\leadsto$ $\lnot\ex{Block}(b)$ \\ implicit effect: remove any $\ex{on}(b,y)$, $\ex{on\_table}(b,y)$, $\ex{on\_block}(b,y)$ %, $\ex{has\_color}(b,c)$

Implicitly, we know that $b_2$ is blocked ($\ex{Blocked}(b_2)$) since $b_1$ is on $b_2$ ($\ex{on\_block}(b_1,b_2)$) and every block that has another block on top is blocked ($\exists\ex{on\_block}^-\sqsubseteq\ex{Blocked}$).
On the other hand, we know that $\ex{on\_block}(b_1,b_3)$ cannot hold, since the $\ex{on\_block}$ relation is functional ($\funct \ex{on\_block}$).

Consider now the action $\ex{move}(x,y,z)$ that moves Block~$x$ from position~$y$ to~$z$.
Its precondition is $[\ex{on}(x,y)]\land\lnot[\ex{Blocked}(x)]\land\lnot[\ex{Blocked}(z)]$, where the atoms in brackets are evaluated \wrt the ontology axioms (epistemic semantics).
For example, the action is applicable for the substitution $x\mapsto b_1$, $y\mapsto b_2$, $z\mapsto b_3$, since $\ex{on\_block}$ is included in $\ex{on}$ and neither $\ex{Blocked}(b_1)$ nor $\ex{Blocked}(b_3)$ are entailed.
\end{example}
However, one property of this formalism is that action effects operate directly on the state, ignoring implicit knowledge, and only check whether the subsequent state is still consistent with the TBox.

\begin{example}\label{ex:indirect-effects}
The effect of the ground action $\ex{move}(b_1,b_2,b_3)$ is to add $\ex{on\_block}(b_1,b_3)$ to the state. % (or $\ex{on\_table}(b_1,b_3)$ if $b_3$ was a $\ex{Table}$).
In the eKAB formalism, this would make the state inconsistent, as argued above.

To obtain a consistent state, we could remove $\ex{on}(b_1,b_2)$.
However, since this fact is not explicitly present in the state (ABox), this operation would not affect the state at all and $[\ex{on}(b_1,b_2)]$ would continue to hold due to $\ex{on\_block}(b_1,b_2)$.

Moreover, even if we explicitly remove $\ex{on\_block}(b_1,b_2)$, we would lose the information that $b_2$ is a block, which means that we should add $\ex{Block}(b_2)$ as well.
\end{example}
%
% In Example \ref{example:cat}, Hanzo is currently a kitten and hence is a cat and an animal (but not a mouse). Assume that he grew up and is no longer a kitten, the corresponding eKAB update would simply delete the fact $Kitten("Hanzo")$ from the current state (ABox). However, this update would also indirectly remove the fact that Hanzo is a cat (and an animal resp.) in the subsequent state. 
%
This illustrates that actions can cause three types of implicit effects: removing a fact requires \tagtext{effect:delete}{(i)} removing all stronger facts and \tagtext{effect:closure}{(ii)} adding previously implied facts to avoid losing information, whereas adding a fact requires \tagtext{effect:conflict}{(iii)} removing any conflicting facts to ensure consistency.

Addressing these problems, \inlinecite{DBLP:journals/jair/GiacomoORS21} introduced the \emph{coherence update semantics} for updating an ABox in the presence of a \DLLite TBox.
In our example, adding $\ex{on\_block}(b_1,b_3)$ would automatically remove $\ex{on\_block}(b_1,b_2)$~\linktext{effect:conflict} and add $\ex{Block}(b_2)$~\linktext{effect:closure}.
Similarly, removing $\ex{on}(b_1,b_2)$ would also remove $\ex{on\_block}(b_1,b_2)$~\linktext{effect:delete}.
Notably, the updated ABox can be computed with the help of a non-recursive Datalog$^\neg$ program.

\inlinecite{DBLP:journals/jair/GiacomoORS21} only considered single-step ABox updates.
However, for planning, such implicit effects need to be taken into account for each action on the way to the goal.
In this paper, we extend eKAB planning by applying the coherence update semantics to action effects.
We investigate the complexity of the resulting formalism of \emph{ceKABs (coherent eKABs)} and show that it is not higher than for classical planning.
In fact, ceKAB planning tasks can be compiled into PDDL with \emph{derived predicates} by utilizing Datalog$^\neg$ programs describing eKAB \cite{DBLP:conf/aaai/Borgwardt0KKNS22} and coherence semantics \cite{DBLP:journals/jair/GiacomoORS21}.
Moreover, we evaluate the feasibility of our approach in off-the-shelf planning systems and the overhead incurred compared to the original eKAB semantics.

Full proofs of all results as well as the implementation and benchmarks are available in the supplementary material.

\subsection{Closely Related Work}

\inlinecite{DBLP:journals/ai/LiuLMW11} formalized instance-level updates for expressive DLs where the update result can be expressed in the same description logic.
\inlinecite{DBLP:conf/aaai/AhmetajCOS14} described integrity constraints and states of graph-structured data over an action language where actions insert and delete nodes/labels using expressive DLs.
The eKAB formalism was optimized and extended to support all Datalog$^\neg$-rewritable Horn DLs, via a compilation into PDDL with derived predicates \cite{DBLP:conf/kr/Borgwardt0KS21,DBLP:conf/aaai/Borgwardt0KKNS22}.
A similar approach uses a black-box, justification-based algorithm to compile an ontology-mediated planning problem into classical planning, even for non-Horn DLs \cite{DBLP:conf/ecai/JohnK24}.

% \stefan{discuss literature: other update semantics proposed for DLs, planning+ontology paper by Koopmann and John; anything from the planning side?}

%% file: parts/pre.tex
\section{Preliminaries}

We introduce all relevant formalisms, including \DLLite, PDDL, eKABs, and the coherence update semantics. For more details, we refer to the original papers.

\subsection{\texorpdfstring{The Description Logic $\DLLite_{core}^{(\mH\mF)}$}{The Description Logic DL-Lite-core-HF}}\label{sec:dllite}

We consider $\DLLite_{core}^{(\mH\mF)}$ \cite{DBLP:journals/jair/ArtaleCKZ09}, which we simply call \DLLite in the following.\footnote{\inlinecite{DBLP:conf/ijcai/CalvaneseMPS16} and \inlinecite{DBLP:journals/jair/GiacomoORS21} used $\DLLite_A$, which is $\DLLite_{core}^{(\mH\mF)}$ extended with attributes.}
Given disjoint sets $\NC,\NR,\NI$ of \emph{atomic concepts}~$A$, \emph{atomic roles}~$P$, and \emph{individuals} (\emph{constants})~$c$,
\emph{(general) concepts and roles} are formed as follows:
\begin{align*}
    \textit{basic role: }&Q \longrightarrow P \mid P^- 
    &\textit{role: }&R \longrightarrow Q \mid \lnot Q \\
    \textit{basic concept: }&B \longrightarrow A \mid \exists Q 
    &\textit{concept: }&C \longrightarrow B \mid \lnot B
\end{align*}
For the \emph{inverse role} $P^-$, we set $(P^-)^- := P$.
A \emph{TBox} (a.k.a.\ \emph{ontology}) is a finite set of \emph{concept inclusions} $B \sqsubseteq C$, \emph{role inclusions} $Q \sqsubseteq R$, and \emph{functionality axioms} $(\texttt{funct}\ Q)$, where an inclusion is called \emph{negative} if it contains $\lnot$, and \emph{positive} otherwise, and functional basic roles and their inverses are not allowed to occur on the right-hand side of any positive inclusions.
An \emph{ABox} is a finite set of \emph{concept assertions} $A(c)$ and \emph{role assertions} $P(c,c')$, where $c,c'$ are individuals.
A \emph{knowledge base} (\emph{KB}) $\mathcal{K}$ is a tuple $\langle \mT, \mA \rangle$, where $\mT$ is a TBox and $\mA$ is an ABox.
The \emph{signature} of \mK, \mA, or \mT is the subset of symbols $\NC\cup\NR\cup\NI$ used in it.

An \emph{interpretation} $\mI=(\Delta^{\mI}, \cdot^{\mI})$ consists of a non-empty set $\Delta^{\mI} \supseteq \NI$ (the \emph{domain} of objects under the \emph{standard name assumption}) and an \emph{interpretation function} $\cdot^{\mI}$ that maps atomic concepts $A$ to subsets $A^{\mI} \subseteq \Delta^{\mI}$ and roles $P$ to relations $P^{\mI} \subseteq (\Delta^{\mI})^2$.
This function is extended to
\begin{align*}
    &(P^-)^{\mI} = \{(o',o) \mid (o,o') \in P^{\mI}\}
    &&(\lnot Q)^{\mI} = (\Delta^{\mI})^2 \setminus Q^{\mI} \\
    &(\exists Q)^{\mI} = \{o \mid \exists o'.\, (o,o') \in Q^{\mI}\}
    &&(\lnot C)^{\mI} = \Delta^{\mI} \setminus C^{\mI}
\end{align*}

An interpretation $\mI$ \emph{satisfies} $B \sqsubseteq C$ if $B^{\mI} \subseteq C^{\mI}$; $Q \sqsubseteq R$ if $Q^{\mI} \subseteq R^{\mI}$; $\texttt{funct}\ Q$ if $(o, o'),(o,o'') \in Q^{\mI}$ implies $o'=o''$; $A(c)$ if $c \in A^{\mI}$; and $P(c,c')$ if $(c, c') \in P^{\mI}$. It is a \emph{model} of a knowledge base $\mathcal{K}$ if $\mI$ satisfies all axioms in~\mK. If every model of $\mathcal{K}$ satisfies an axiom $\alpha$, then $\mathcal{K}$ \emph{entails} $\alpha$ (written $\mathcal{K} \models \alpha$). 
\mK is \emph{consistent} if it has a model, and an ABox \mA is \emph{consistent with} a TBox~\mT if $\langle \mT, \mA \rangle$ is consistent.
% For a KB $\mathcal{K}$, a concept $B$ is \emph{subsumed} by a concept $C$ w.r.t.\ $\mathcal{K}$ if $\mathcal{K} \models B \sqsubseteq C$. Finally, a concept $C$ (or role $R$) is \emph{satisfiable} w.r.t.\ $\mathcal{K}$ if there is a model $\mI$ of $\mathcal{K}$ where $C^{\mI} \neq \emptyset$ (or $R \neq \emptyset$).
% 

% Let $\mT$ be a TBox expressed in a DL $\mathcal{L}$.
The \emph{$\mathcal{T}$-closure} $cl_{\mathcal{T}}(\mA)$ of $\mA$ \wrt $\mathcal{T}$ is the set of all assertions over the signature of $\langle \mT, \mA \rangle$ that are entailed by $\langle \mT, \mA \rangle$.
Two ABoxes $\mA$ and $\mA^\prime$ are \emph{equivalent} w.r.t.\ $\mathcal{T}$ if $cl_{\mathcal{T}}(\mA) = cl_{\mathcal{T}}(\mA^\prime)$.
The \emph{deductive closure} $cl(\mT)$ of $\mT$ is the set of all TBox axioms over the signature of~\mT that are entailed by~$\mT$, and can be computed in polynomial time \cite{DBLP:journals/jar/CalvaneseGLLR07,DBLP:journals/jair/ArtaleCKZ09}.

\begin{example}
    For Example~\ref{ex:blocks}, we have
    \begin{align*}
        \mT &= \{\ex{on\_block} \sqsubseteq \ex{on},\ \exists \ex{on\_block}^- \sqsubseteq \ex{Block},\ \dots \} \\
        \mA &= \{\ex{on\_block}(b_1,b_2),\ \ex{on\_table}(b_3,t)\} \\
        {cl}_\mT(\mA) &= \{\ex{on}(b_1,b_2),\ \ex{Block}(b_1),\ \ex{Block}(b_2),\ \dots \} \\
        {cl}(\mT) &= \{\exists\ex{on\_block}\sqsubseteq\ex{Block},\ \exists\ex{on}\sqsubseteq\lnot\ex{Table},\ \dots \}
    \end{align*}
\end{example}

\subsubsection{Queries and Datalog}
A \emph{conjunctive query} (\emph{CQ}) is an FO-formula of the form $q(\Vec{x}) \coloneqq \exists \Vec{y}.\Phi(\Vec{x}, \Vec{y})$ where $\Phi$ is a conjunction atoms of the form $A(x)$ or $P(x,y)$. A \emph{union of conjunctive queries} (UCQ) is a disjunction of CQs with the same free variables.
Given a KB $\langle\mT,\mA\rangle$ and a substitution~$\theta$ of the variables in~$\vec{x}$ by constants from~\mA,
the \emph{UCQ answering} problem is to decide whether $\langle\mT,\mA\rangle$ entails the ground UCQ $\theta(q)$, which we denote as $\mA, \mT, \theta \models q$.
Assuming \wlg that \mT contains two disjoint atomic concepts $A\sqsubseteq\lnot B$, the special CQ $\bot:=\exists x.A(x)\land B(x)$ can be used to check consistency of $\langle\mA,\mT\rangle$.

In the following, we use \emph{states}~$s$ instead of ABoxes, which are finite sets of ground atoms (\emph{facts}) $p(\vec{c})$ over a finite set $\mP$ of predicates of arbitrary arity.
When reasoning in \DLLite, we only consider the unary and binary atoms in a state.
Further, we denote by $\NI(s)$ the constants occurring in~$s$.
\inlinecite{DBLP:conf/ijcai/CalvaneseGLLR07a} introduced \emph{extended conjunctive queries} (ECQs) to combine open- and closed-world reasoning.
An ECQ $Q$ is constructed as follows:
\[ Q \coloneqq p(\Vec{x}) \ |\ [q(\vec{x})] \ |\ \lnot Q \ |\ Q_1 \land Q_2 \ |\ \exists y.\,Q \,, \]
where $p$ is a predicate (of any arity), $\Vec{x}$ are terms (constants or variables) and $q(\vec{x})$ is a UCQ.
Intuitively, $[q(\vec{x})]$ denotes the evaluation of~$q(\vec{x})$ w.r.t.\ a KB $\langle\mathcal{T},s\rangle$, whereas $q(\vec{x})$ (without brackets $[\cdot]$) is evaluated directly over the minimal model of the state~$s$.
The complete semantics of ECQs is as follows:
\begin{align*}
    &s, \mT, \theta \models p(\Vec{x}) & \ \text{iff} & \ s \models p(\theta(\Vec{x}))\\
    &s, \mT, \theta \models [q] & \ \text{iff} & \ s,\mT,\theta \models q\\
    &s, \mT, \theta \models \lnot Q & \ \text{iff} & \ s,\mT,\theta \not\models Q\\
    &s, \mT, \theta \models Q_1 \land Q_2 & \text{iff} & \ s,\mT,\theta \models Q_1\ \text{and} \ s,\mT,\theta \models Q_2 \\
    &s, \mT, \theta \models \exists y.Q  & \text{iff} & \ \exists d \in \NI(s).\; s,\mT, \theta[y \rightarrow d] \models Q
\end{align*}

Except for $[q]$, this corresponds to the standard (closed-world) evaluation of FO-formulas. 

A \emph{Datalog}$^\neg$ \emph{rule} \cite{DBLP:books/aw/AbiteboulHV95} has the form $p(\Vec{x}) \leftarrow \Phi(\Vec{x}, \Vec{y})$, where the \emph{head} $p(\Vec{x})$ is an atom and the \emph{body} $\Phi(\Vec{x}, \Vec{y})$ is a conjunction of \emph{literals}, \ie atoms or negated atoms. A finite set of Datalog$^\neg$ rules (\emph{program}) $\mR$ is \emph{stratified} if there is a partition $\mathcal{P}_1, \dots, \mathcal{P}_n$ of the set of predicates $\mathcal{P}$ appearing in $\mR$ s.t.\ for each predicate $p_i \in \mathcal{P}_i$ and $p_i(\Vec{x}) \leftarrow \Phi(\Vec{x}, \Vec{y}) \in \mR$ ($i \in \{1, \dots, n\}$), the following holds: (1) If $p_j \in \mathcal{P}_j$ appears in $\Phi(\Vec{x}, \Vec{y})$, then $j \leq i$; (2) If $\lnot p_j \in \mathcal{P}_j$ appears in $\Phi(\Vec{x}, \Vec{y})$, then $j < i$.
We consider only stratified Datalog$^\neg$ programs.
 
For a state~$s$ and a program~$\mR$, $\mR(s)$ denotes the minimal Herbrand model of $s \cup \mR$, where all variables are implicitly unversally quantified.
% We consider Datalog rewritability similar to the work of \citeauthor{DBLP:conf/aaai/Borgwardt0KKNS22} 2022. 
Let $\mT$ be a TBox and $q(\Vec{x})$ be an UCQ. Then, $\mT$ and $q(\Vec{x})$ are \emph{Datalog$^\neg$-rewritable} if there is a Datalog$^\neg$ program $\mR_{\mT,q}$ and a predicate $P_q$ s.t., for every state $s$ and every mapping $\theta$ of $\Vec{x}$ to $\NI(s)$, it holds that $s, \mT, \theta \models q(\vec{x})$ iff $\mR_{\mT,q}(s) \models P_q(\theta(\Vec{x}))$. 
When using \DLLite TBoxes, all UCQs can be rewritten in this way \cite{DBLP:journals/jar/CalvaneseGLLR07,DBLP:conf/aaai/EiterOSTX12}, even without negation or recursion.
Moreover, \inlinecite{DBLP:conf/ijcai/CalvaneseGLLR07a} extended rewritability to ECQs as follows:

\begin{proposition}\label{prop:calvanese}
    Given an ECQ $Q$, let $\mR_{\mT,Q}$ be the disjoint union of $\mR_{\mT,q}$ for all $[q(\vec{x})]$ in~$Q$, and let the FO-formula $Q_{\mT}$ be the result of substituting each $[q(\Vec{x})]$ in $Q$ with $P_q(\Vec{x})$. Then, $s, \mT, \theta \models Q(\Vec{x})$ iff $\mR_{\mT,Q}(s) \models Q_{\mT}(\theta(\Vec{x}))$, for all $s,\theta$.
\end{proposition}

\begin{example}
    In the blocks world example, the query $\bot$ can be rewritten into the following rules, among others:
\begin{align*}
  P_\bot &\leftarrow \ex{on\_block}(x,y), \ex{on\_table}(x,z) \\
  P_\bot &\leftarrow P_{\ex{Block}}(y), P_{\ex{Table}}(y) \\
  P_\bot &\leftarrow \ex{on\_block}(x,y), \ex{on\_block}(x,z), y\neq z \\
  % P_{\exists\ex{on\_block}}(x) &\leftarrow \exists y.\,\ex{on\_block}(x,y) \\
  % P_{\exists\ex{on\_table}}(x) &\leftarrow \exists y.\,\ex{on\_table}(x,y) \\
  P_{\ex{Block}}(x) &\leftarrow \ex{Block}(x) \\
  P_{\ex{Block}}(x) &\leftarrow \ex{on\_block}(x,y) \\
  P_{\ex{Block}}(x) &\leftarrow P_{\ex{on}}(x,y)
\end{align*}
The ECQ $[\ex{on}(x,y)]\land\lnot[\ex{Blocked}(x)]\land\lnot[\ex{Blocked}(z)]$ can be rewritten to $P_{\ex{on}}(x,y)\land\lnot P_{\ex{Blocked}}(x)\land\lnot P_{\ex{Blocked}}(z)$.
\end{example}

\subsection{PDDL}\label{sec:pddl}
We use PDDL~2.1 \cite{DBLP:journals/jair/FoxL03} with stratified derived predicates. Here, states are viewed under the closed-world assumption, i.e.\ atoms absent from a state $s$ are treated as false.  
Let $\mP_{der}$ be a finite set of \emph{derived predicates}.
In the following, all FO-formulas are constructed from $\mathcal{P} \cup \mathcal{P}_{der}$ unless explicitly stated otherwise.
However, states only contain facts over \mP, and hence action effects cannot add or remove facts over $\mP_{der}$.
An \emph{action} is a tuple $(\Vec{x}, \texttt{pre}, \texttt{eff})$ consisting of parameters, a precondition, and a finite set of effects. A \emph{precondition} is an FO-formula with free variables from $\Vec{x}$ and a \emph{(conditional) effect} is a tuple $(\Vec{y}, \texttt{cond}, \texttt{add}, \texttt{del})$. Here, $\Vec{y}$ are additional variables, \texttt{cond} is an FO-formula with free variables from $\Vec{x} \cup \Vec{y}$, and \texttt{add} and \texttt{del} are finite sets of atoms and negated atoms, respectively, containing free variables from $\Vec{x} \cup \Vec{y}$ and predicates from~$\mathcal{P}$.

A PDDL \emph{domain description} is a tuple $(\mathcal{P}, \mathcal{P}_{der}, \mathcal{A}, \mathcal{R})$, where $\mathcal{A}$ is a finite set of actions and $\mathcal{R}$ is a finite set of FO-rules of the form $p(\Vec{x}) \leftarrow \Phi(\Vec{x})$, where $p \in \mathcal{P}_{der}$ and $\Phi$ is an FO-formula. 
The set $\mathcal{R}$ is also restricted to be \emph{stratified}, which is defined similarly as for $\text{Datalog}^{\neg}$ programs by restricting the positive and negative occurrences of predicates in $\Phi(\vec{x})$ depending on the location of~$p$ in a suitable partition of~\mP.
While not every stratified set of FO-rules has an equivalent stratified Datalog$^\neg$ program \cite{RoegerG-PuK24}, the converse trivially holds.
A PDDL \emph{task} is a tuple $(\Delta, \mathcal{O}, \mathcal{I}, \mG)$, where $\Delta$ is a PDDL domain description, $\mathcal{O}$ is a finite set of constants including those used in~$\Delta$, $\mathcal{I}$ is an \emph{initial state}, and $\mG$ is the \emph{goal}, a closed FO-formula.
Here, $\mO$ plays the same role as the set $\NI$ for DLs, and for consistency we will use the notation $\mO$ in both settings from now on.

Substituting the parameters in $a = (\Vec{x}, \texttt{pre}, \texttt{eff})$ according to an assignment $\theta$ from $\Vec{x}$ to constants in $\mathcal{O}$ yields the \emph{ground} action $\theta(a)$.
For a ground action, we omit $\Vec{x}$ and write $(\texttt{pre}, \texttt{eff})$. 
Given a state $s$, the ground action $a = (\texttt{pre}, \texttt{eff})$ is \emph{applicable in~$s$} if $\mathcal{R}(s) \models \texttt{pre}$, which is defined similarly to Datalog$^\neg$.
The \emph{application} of $a$ to $s$ then produces a new state $s \llbracket a \rrbracket$ that contains a fact $\alpha$ iff either
\begin{itemize}
    \item there is $(\Vec{y}, \texttt{cond}, \texttt{add}, \texttt{del}) \in \texttt{eff}$ and an assignment $\theta$ such that $\mathcal{R}(s) \models \theta(\texttt{cond})$ and $\alpha \in \theta(\texttt{add})$, or
    \item $\alpha \in s$ and, for all $(\Vec{y}, \texttt{cond}, \texttt{add}, \texttt{del}) \in \texttt{eff}$ and $\theta$ with $\mathcal{R}(s) \models \theta(\texttt{cond})$, it holds that $\neg\alpha \not\in \theta(\texttt{del})$.
\end{itemize} 

Intuitively, an effect inserts and removes facts according to $\theta(\texttt{add})$ and $\theta(\texttt{del})$ whenever $\theta(\texttt{cond})$ is true in $\mR(s)$.
If an action concurrently inserts and deletes $\alpha$ due to conflicting effects, the insertion takes precedence by definition.

A sequence of ground actions $\Pi = a_1, \dots, a_n$ yields a sequence of states $s$, $s\llbracket a_1\rrbracket$, $s\llbracket a_1\rrbracket\llbracket a_2 \rrbracket$, \dots, $s\llbracket a_1\rrbracket\dots\llbracket a_n\rrbracket$ if each $a_i$ is applicable in the preceding state, in which case $\Pi$ is called \emph{applicable in $s$}.
$\Pi$ is a \emph{plan} for the PDDL task $(\Delta, \mathcal{O}, \mathcal{I}, \mG)$ if $\Pi$ is applicable in $\mathcal{I}$ and $\mathcal{R}(\mathcal{I} \llbracket \Pi \rrbracket) \models \mG$.
The \emph{length} of this plan is $\|\Pi\|:=n$.

Although derived predicates are often implemented by treating them as simple actions that can be applied after each normal action, they can be used to encode planning problems more succinctly.
In general, derived predicates cannot be compiled away without increasing either the size of the PDDL task or the length of the plans super-polynomially \cite{DBLP:journals/ai/ThiebauxHN05}.

\subsection{eKABs}\label{sec:ekabs}

An \emph{explicit-input knowledge and action base (eKAB) domain description} \cite{DBLP:conf/ijcai/CalvaneseMPS16} is a tuple $(\mathcal{P}, \mathcal{A}, \mathcal{T})$, where $\mathcal{P}$ is a finite set of predicates, $\mathcal{A}$ a finite set of actions that use ECQs instead of FO-formulas, and $\mathcal{T}$ a TBox over the unary and binary predicates in $\mathcal{P}$.
An \emph{eKAB task} is a tuple $(\Delta, \mathcal{O}, \mathcal{O}_0, \mathcal{I}, \mG)$, where $\Delta$ is an eKAB domain description, $\mO$ is a set of constants (potentially infinite), $\mO_{0}$ is a finite subset of $\mO$ that includes all constants used in $\Delta$, the initial state $\mathcal{I}$ uses only constants from~$\mathcal{O}_0$ and is consistent with $\mathcal{T}$, and the goal $\mG$ is a closed ECQ using only constants from $\mathcal{O}_0$.
% 

% Similar to PDDL, we also omit $\Vec{x}$ for a ground action $a = (\texttt{pre}, \texttt{eff})$ resulting from the mapping $\theta\colon \Vec{x} \rightarrow \mO$.
A ground action $a$ is \emph{applicable} in state $s$ if $s, \mathcal{T} \models \texttt{pre}$ and $\langle\mT,s \llbracket a \rrbracket\rangle$ is consistent, where $s \llbracket a \rrbracket$ is defined as for PDDL, but replacing $\mathcal{R}(s) \models \theta(\texttt{cond})$ by $s, \mathcal{T}, \theta \models \texttt{cond}$.
Plans~$\Pi$ are also defined as before, but the goal condition is now expressed as $\mI\llbracket \Pi \rrbracket, \mT \models \mG$.
We use the same notation $\llbracket\cdot\rrbracket$ as for eKABs, since it is usually clear from the context which definition we are referring to.

As usual, we assume that all eKABs are \emph{state-bounded}, \ie there is a bound $b\in\mathbb{N}$ s.t.\ any state reachable from  $\mI$ contains at most $b$ objects, to obtain manageable state transition systems \cite{DBLP:conf/aiia/CalvaneseGMP13,DBLP:conf/atal/GiacomoLPV14}.
For simplicity, we can then assume that $\mO = \mO_{0}$ and denote both as $\mO$.
Additionally, as in \cite{DBLP:conf/aaai/Borgwardt0KKNS22}, we assume \wlg that the goal $\mG$ consists of a single ground atom $g(\vec{c})$ (without brackets $[\cdot]$).

\begin{example}\label{ex:ekab}
    The $\ex{move}$ action from the introduction consists of the parameters $(x,y,z)$, precondition $[\ex{on}(x,y)]\land\lnot[\ex{Blocked}(x)]\land\lnot[\ex{Blocked}(z)]$ and effects
    \begin{align*}
        & ((),\ [\ex{Block}(y)],\ \emptyset,\ \{\lnot\ex{on\_block}(x,y)\}), \\
        & ((),\ [\ex{Table}(y)],\ \emptyset,\ \{\lnot\ex{on\_table}(x,y)\}), \\
        & ((),\ [\ex{Block}(z)],\ \{\ex{on\_block}(x,z)\},\ \emptyset), \\
        & ((),\ [\ex{Table}(z)],\ \{\ex{on\_table}(x,z)\},\ \emptyset),
    \end{align*}
    which essentially remove $\ex{on}(x,y)$ and add $\ex{on}(x,z)$. %, depending on the nature of~$y$ and~$z$, respectively.
\end{example}

\subsubsection{eKAB-to-PDDL Compilation} A \emph{compilation scheme} $\textbf{f}$ \cite{DBLP:journals/ai/ThiebauxHN05,DBLP:conf/aaai/Borgwardt0KKNS22} is a tuple of functions $(f_{\delta}, f_o, f_i, f_g)$ that translates an eKAB task $\mE = (\Delta, \mO, \mI, \mG)$ to a PDDL task $F(\mE) \coloneqq (f_{\delta}(\Delta), \mO \cup f_o(\Delta), f_i(\mO, \mI), f_g(\mO,\mG))$ s.t.\ a plan for $\mE$ exists iff a plan for $F(\mE)$ exists, and $f_i$, $f_g$ are computable in polynomial time.
If the size of $f_{\delta}(\Delta)$ and $f_o(\Delta)$ is polynomial w.r.t.\ the size of $\Delta$, we say that \textbf{f} is \emph{polynomial}. If, or every plan $\Pi$ for $\mE$, there is a plan $\Pi'$ for $F(\mE)$ s.t.\ $\norm{\Pi'} \leq c \cdot \norm{\Pi}^n + k$ where $c,n,k$ are positive integers, then \textbf{f} \emph{preserves plan size polynomially}. If $n=1$ (and $c=1$), then \textbf{f} preserves plan size \emph{linearly} (\emph{exactly}).

A state-bounded eKAB task $\mE = ((\mP, \mA, \mT), \mO, \mI, \mG)$ can be compiled into PDDL by viewing the Datalog$^{\neg}$-rewritings of all ECQs as FO-rules and additionally checking consistency via the rewriting of~$\bot$.\footnote{This is a simplified version of the compilation from \inlinecite{DBLP:conf/aaai/Borgwardt0KKNS22} that does not allow fresh objects in rewritings.}
\begin{itemize}
\item Construct a copy $\mT'$ of \mT and copies $Q'$ of all ECQs~$Q$ in~\mE by replacing each predicate $p$ by $p'$.
\item Define $\mR^\prime := \mR_{\mT^\prime} \cup \{ p^\prime(\Vec{x}) \leftarrow p(\Vec{x}) \mid p \in \mP \}$, where $\mR_{\mT^{\prime}}$ is the disjoint union of $\mR_{\mT^\prime, \bot}$ and all $\mR_{\mT^\prime, Q'}$ for ECQs~$Q$ in~\mE (cf.\ Proposition~\ref{prop:calvanese}).
\item Define $F(\mE):=((\mP, \mP^{\prime}, \mA^{\prime}, \mR^{\prime}), \mO, \mI, \mG^{\prime})$, where $\mP^\prime$ contains all predicates from $\mR_{\mT^\prime}$, $\mG^\prime:=\neg P_{\bot} \land \mG$, and $\mA^\prime$ is obtained from $\mA$ by replacing all preconditions \texttt{pre} by $\neg P_{\bot} \land \texttt{pre}^\prime_{\mT^\prime}$ and effect conditions \texttt{cond} by $\texttt{cond}^\prime_{\mT^\prime}$.
\end{itemize}
The condition $\lnot P_\bot$ checks consistency w.r.t.~\mT, as required by the semantics.
For \DLLite TBoxes, this is a polynomial compilation scheme that preserves plan size exactly \cite{DBLP:conf/ijcai/CalvaneseGLLR07a,DBLP:conf/aaai/EiterOSTX12,DBLP:conf/aaai/Borgwardt0KKNS22}.

\subsection{Instance-Level Coherence Update}\label{sec:cus}

An \emph{(instance-level) update} for \DLLite \cite{DBLP:journals/jair/GiacomoORS21} is a set $\mU$ consisting of \emph{insertions} $\textit{ins}(P(\Vec{c}))$ and \emph{deletions} $\textit{del}(P(\Vec{c}))$, where $P(\Vec{c})$ is an ABox assertion.
For an update $\mathcal{U}$, the set of ABox assertions occurring in insertions (deletions) in $\mathcal{U}$ is denoted $A^{+}_{\mathcal{U}}$ ($A^{-}_{\mathcal{U}}$).
Let $\mK = \langle \mathcal{T}, \mA \rangle$ be a consistent \DLLite ontology and $\mU$ an update.
\begin{definition}\label{def:update_semantics}
    An ABox $\mA^\prime$ \emph{accomplishes the update} of $\mathcal{K}$ with $\mathcal{U}$ if $\mA^\prime = \mA^{\prime\prime} \cup A^{+}_{\mathcal{U}}$ for some maximal subset $\mA^{\prime\prime}\subseteq\mA$ s.t.\ $\mA^\prime$ is consistent with $\mathcal{T}$ and $\langle \mathcal{T}, {\mA}^\prime  \rangle \not\models \beta$ for all $\beta \in A^{-}_{\mathcal{U}}$.
\end{definition}
Intuitively, $\mA'$ should differ from \mA as little as possible.
Moreover, \mU cannot add and delete the same assertion (explicitly or implicitly), as described by the following result.

\begin{proposition}[\citeauthor{DBLP:journals/jair/GiacomoORS21} \citeyear{DBLP:journals/jair/GiacomoORS21}]\label{pro:existence_of_abox}
    An ABox $\mA'$ accomplishing the update of $\mK$ with $\mU$ exists iff $A^{+}_{\mU}$ is consistent with $\mT$ and $cl_{\mT}(A^{+}_{\mU}) \cap A^{-}_{\mU} = \emptyset$.
    Then, $\mA'$ is unique.
\end{proposition}
$\mU$ is \emph{compatible with $\mT$} if it satisfies these two conditions, which are independent of the ABoxes~$\mA,\mA'$.
To make the semantics independent of the syntax of \mA, the following definition considers the $\mT$-closure $cl_{\mT}(\mA)$ instead of \mA itself.

\begin{definition}[Coherence Update Semantics]\label{def:cus}
    If $\mathcal{U}$ is compatible with $\mathcal{T}$, the \emph{result $\mK \bullet \mU$ of updating $\mathcal{K}$ with $\mathcal{U}$} is $\langle \mathcal{T}, {\mA}^{\prime\prime} \rangle$, where ${\mA}^{\prime\prime}$ is equivalent (w.r.t.\ $\mathcal{T}$) to the ABox ${\mA}^{\prime}$ that accomplishes the update of $\langle \mathcal{T}, cl_{\mathcal{T}}({\mA}) \rangle$ with $\mathcal{U}$.
\end{definition}
Since ${\mA}''$ is unique up to equivalence w.r.t.~\mT, $\mK \bullet \mU$ is treated as unique in the following.

\citeauthor{DBLP:journals/jair/GiacomoORS21} (\citeyear{DBLP:journals/jair/GiacomoORS21}) construct a Datalog$^\neg$ program $\mR_\mT^{\mathsf{u}}$ to compute the effects of an update w.r.t.~\mT.
For this, $\mA$ and~\mU are encoded into a single dataset $D_{\mU,\mA}$ that contains all assertions from \mA as well as $\textit{ins\_p\_request}(\Vec{c})$ ($\textit{del\_p\_request}(\Vec{c})$) for each $\textit{ins}(P(\Vec{c}))$ ($\textit{del}(P(\Vec{c}))$) in $\mU$.
To check whether $\mU$ is compatible with $\mT$, $\mR_\mT^{\mathsf{u}}$ uses the predicate $\textit{incompatible\_update}()$ that is derived when the conditions of Proposition~\ref{pro:existence_of_abox} are violated.
%
% For each concept or role $P$ appearing in $\mK$,
The derived predicates
% $\textit{ins\_p\_request}(\Vec{x})$ and $\textit{del\_p\_request}(\Vec{x})$ encode the request for insertion/deletion of elements of $P$ according to~\mU,
% $\textit{ins\_p\_closure}(\Vec{x})$ indicates the insertion of elements into $P$ due to the closure $cl_{\mT}(\mA)$, and
$\textit{ins\_p}(\Vec{x})$ and $\textit{del\_p}(\Vec{x})$ represent the final insertion and deletion operations needed to obtain $\mK \bullet \mU$ from~$\mK$.

This construction is correct in the sense that the resulting ABox $\mA_{\mU,\mT}$ is equivalent to $\mA''$ from Definition~\ref{def:cus}, where $\mA_{\mU,\mT}$ is obtained from $\mA$ by inserting (deleting) an assertion $P(\Vec{o})$ iff $\mR_\mT^{\mathsf{u}}(D_{\mU, \mA}) \models \textit{ins\_p}(\Vec{o})$ ($\mR_\mT^{\mathsf{u}}(D_{\mU, \mA}) \models \textit{del\_p}(\Vec{o})$).

\begin{proposition}[\citeauthor{DBLP:journals/jair/GiacomoORS21} \citeyear{DBLP:journals/jair/GiacomoORS21}]\label{pro:coherence_update_semantics_correctness}
    \mU is compatible with \mT iff $\mR^{\mathsf{u}}_\mT(D_{\mU,\mA})\not\models\textit{incompatible\_update}()$.
    In this case, for all assertions $P(\Vec{o})$ over the signature of $\langle\mT,\mA\rangle$, we have $\langle \mT, \mA_{\mU,\mT} \rangle \models P(\Vec{o})$ iff $\mK \bullet \mU \models P(\Vec{o})$.
\end{proposition}

% \stefan{Maybe shorten the example below? It is quite long.}

\begin{example}
Following Example~\ref{ex:ekab}, we express the effect of $\ex{move}(b_1, b_2, b_3)$ by the update $\mU=\{ \textit{del}(\ex{on\_block}(b_1, b_2)),$ $\textit{ins}(\ex{on\_block}(b_1,b_3)) \}$.
Using coherence update semantics, we do not have to distinguish the type of~$b_2$ and can simply use $\textit{del}(\ex{on}(b_1,b_2))$ instead, which yields the facts
    $\ex{on\_block}(b_1,b_2)$, $\ex{on\_table}(b_3,t)$, 
    $\textit{del\_on\_request}(b_1,b_2)$ and
    $\textit{ins\_on\_block\_request}(b_1,b_3)$.
    % \mA \cup \{ \textit{del\_on\_request}(b_1,b_2), 
    % \\ & \ \ \ \ \ \ \ \textit{ins\_on\_block\_request}(b_1,b_3)\} 
% \end{align*}

First, the program $\mR_\mT^{\mathsf{u}}$ translates the requests into direct insertions and deletions:
\begin{align*}
    \textit{del\_on}(x,y) \leftarrow{}
    & \ex{on}(x,y), \textit{del\_on\_request}(x,y) \\
    \textit{ins\_on\_block}(x,y) \leftarrow{}
    & \lnot \ex{on\_block}(x,y), \\
    & \textit{ins\_on\_block\_request}(x,y)
\end{align*}
The first rule has no effect, however, since $\ex{on}(b_1,b_2)$ is not in the ABox.
Instead, we have to remove $\ex{on\_block}(b_1,b_2)$ since $\ex{on\_block}\sqsubseteq\ex{on} \in \mT$ (cf.~\linktext{effect:delete} from Example~\ref{ex:indirect-effects}):
\begin{align*}
    \textit{del\_on\_block}(x,y) \leftarrow{}
    & \ex{on\_block}(x,y), \textit{del\_on\_request}(x,y)
\end{align*}
Additionally, adding $\ex{on\_block}(b_1,b_3)$ also ensures that $\ex{on\_block}(b_1,b_2)$ gets deleted, since otherwise the functionality of $\ex{on\_block}$ would be violated (cf.~\linktext{effect:conflict}):
\begin{align*}
    \textit{del\_on\_block}(x,y) \leftarrow{}
    & \textit{on\_block}(x,y), \\
    & \textit{ins\_on\_block\_request}(x,z), y \neq z
    % \textit{incompatible\_update}() & \leftarrow \textit{ins\_on\_block\_request}(X,Y) 
    % \\ & \land \textit{ins\_on\_block\_request}(X,Z)
    % \\ & \land Y \neq Z
\end{align*}
% 
% Extending the previous grounding to $Z \mapsto b_3$, the above rules derive $\textit{del\_on\_block}(b_1, b_2)$, which indicates the deletion of $\ex{on\_block}(b_1,b_2)$.
Finally, due to $\exists\ex{on\_block}^- \sqsubseteq \ex{Block} \in \mT$, the program retains the information $\ex{Block}(b_2)$ when $\ex{on\_block}(b_1,b_2)$ is deleted, by first deriving $\textit{ins\_block\_closure}(b_2)$ (cf.~\linktext{effect:closure}):
\begin{align*}
    % \textit{del\_on\_block}(X,Y) & \leftarrow \textit{on\_block}(X,Y) 
    % \\ & \land \textit{del\_on\_block\_request}(X,Y)
    % \\
    \textit{ins\_block\_closure}(x) \leftarrow{}
    & \textit{del\_on\_block}(y,x), \lnot \textit{Block}(x), \\
    & \lnot \textit{ins\_block\_request}(x), \\
    & \lnot \textit{del\_block\_request}(x)
\end{align*}
This is then translated into an insertion operation if there are no conflicting requests that would cause an inconsistency (recall that $\ex{Block}\sqsubseteq\lnot\ex{Table}\in\mT$):
\begin{align*}
    \textit{ins\_block}(x) \leftarrow{}
    & \textit{ins\_block\_closure}(x), \lnot\textit{ins\_table\_request}(x)
\end{align*}
In summary, the above rules derive $\textit{ins\_on\_block}(b_1,b_3)$, $\textit{del\_on\_block}(b_1,b_2)$, and $\textit{ins\_block}(b_2)$.

In addition, the program $\mR_\mT^{\mathsf{u}}$ checks the conditions of Proposition~\ref{pro:existence_of_abox}, \eg whether the same tuple is requested to be added to $\ex{on\_block}$ and removed from $\ex{on}$:
\begin{align*}
    \textit{incompatible\_update}() \leftarrow{}
    & \textit{ins\_on\_block\_request}(x,y), \\
    & \textit{del\_on\_request}(x,y)
\end{align*}
\end{example}

However, \inlinecite{DBLP:journals/jair/GiacomoORS21} only considered a single update at a time, whereas we want to determine a sequence of updates to achieve a goal property.

%% file: parts/coherence_planning.tex
\section{eKABs with Coherence Update Semantics}

We now define our new formalism, which we call \emph{coherent eKABs (ceKABs)}.
Syntactically, ceKAB domain descriptions and tasks $((\mP, \mA, \mT), \mO, \mI, \mG)$ are exactly the same as for eKABs.
However, the difference lies in the semantics of the action effects.
First, we define the update $\mU_a$ that is induced by an action~$a$.

\begin{definition}\label{def:update_from_actions}
    Let $a = (\texttt{pre}, \texttt{eff})$ be a ground action and $s$ a state.
    If $s, \mathcal{T} \models \texttt{pre}$, then the \emph{associated update} $\mathcal{U}_a$ is the smallest set s.t.\ for each $(\Vec{y},\texttt{cond}, \texttt{add}, \texttt{del}) \in \texttt{eff}$ and for each assignment $\theta$ with $s, \mathcal{T} \models \theta(\texttt{cond})$:
    \begin{itemize}
        \item for all $\alpha \in \theta(\texttt{add})$, we have $\alpha \in A^{add}_{\mU_a}$; and 
        \item for all $\neg\alpha \in \theta(\texttt{del})$, we have $\alpha \in A^{del}_{\mU_a}$.
    \end{itemize}
    The ground action~$a$ is \emph{applicable in $s$} if $s, \mathcal{T} \models \texttt{pre}$ and $\mU_a$ is compatible with~\mT.
    The \emph{application} of $a$ to $s$ is then the state $s\llbracket a\rrbracket$ for which $\langle\mT,s\rangle\bullet\mU_a=\langle\mT,s\llbracket a\rrbracket\rangle$.
\end{definition}
As before, the operation $\langle\mT,s\rangle\bullet\mU_a$ considers only the unary and binary predicates in~$s$ and ignores any predicates of higher arity.
Moreover, recall that the resulting state $s\llbracket a\rrbracket$ is only unique up to equivalence w.r.t.~\mT (see Definition~\ref{def:cus}).
We again abuse the notation $\llbracket\cdot\rrbracket$ since the semantics we use is usually clear from the context.

In contrast to PDDL and eKABs, for ceKABs, an assertion may be contained in both $A^{add}_{\mU_a}$ and $A^{del}_{\mU_a}$, which could cause $\mU_a$ to be incompatible with~\mT.
Hence, actions need to be specified in such a way to avoid conflicting effects.
Moreover, the compatibility requirement for $\mU_a$ already ensures that $s\llbracket a\rrbracket$ is consistent with~\mT, and thus we do not need to check consistency of $\langle\mT,s \llbracket a \rrbracket\rangle$ as for eKABs.

The definition of $s\llbracket a\rrbracket$ can again be extended to sequences of ground actions as usual.

\begin{definition}\label{def:coherence_plan_ekab}
    A sequence of ground actions $\Pi = a_1, \dots, a_n$ is a \emph{coherence plan} for the ceKAB task $((\mP, \mA, \mT), \mO, \mI, \mG)$ if $\mI\llbracket \Pi \rrbracket, \mT \models \mG$.
\end{definition}

We also define \emph{coherence compilation schemes} as for eKABs, but using coherence plans instead of eKAB plans.

\subsection{Compilation to PDDL}
We now present a polynomial compilation scheme by extending the eKAB-to-PDDL compilation with the update rules $\mR_\mT^{\mathsf{u}}$ and introducing an additional action that implements the effects of each update in the state.
We use the predicate $\textit{updating}()$ to indicate that an update is in progress.
As in Section~\ref{sec:ekabs}, we use the copy~$\mT'$ of~\mT where all predicates are renamed, and correspondingly rename the ECQs~$Q$ occurring in conditions to~$Q'$.

\begin{definition}\label{def:ekab_to_pddl}
    Let $\mathcal{E} = ((\mP, \mA, \mT), \mO, \mI, \mG)$ be a ceKAB task. The PDDL task $F(\mE):=((\mP, \mP^{\prime}_{\mathsf{u}}, \mA^{\prime}_{\mathsf{u}}, \mR^{\prime}_{\mathsf{u}}), \mO, \mI, \mG^{\prime}_{\mathsf{u}})$ is constructed as follows:
    \begin{enumerate}[label=(\arabic*),leftmargin=*]
        \item $\mR^\prime_{\mathsf{u}}$ contains $\mR'$ from Section~\ref{sec:ekabs} and $\mR_\mT^{\mathsf{u}}$ from Section~\ref{sec:cus} as well as, for each predicate $p \in \mP$, the rules
            \begin{flalign*}
                &\textit{updating}()\ \leftarrow \ \textit{ins}\_\textit{p}\_\textit{request}(\Vec{x})\,, \\
                &\textit{updating}()\ \leftarrow \ \textit{del}\_\textit{p}\_\textit{request}(\Vec{x})\,,
            \end{flalign*}
            where $\vec{x}$ matches the arity of~$p$.
        \item $\mP^\prime_{\mathsf{u}}$ contains all predicates from $\mR^\prime_{\mathsf{u}}$.
        \item For each action $a = (\Vec{x}, \texttt{pre}, \texttt{eff}) \in \mA$, $\mA^\prime_{\mathsf{u}}$ contains the \emph{request action} $a^\prime$ obtained by:
        \begin{itemize}
            \item peplacing \texttt{pre} by $\lnot \textit{updating}() \land \texttt{pre}^\prime_{\mT^\prime}$,
            \item for each effect $e = (\Vec{y}, \texttt{cond}, \texttt{add}, \texttt{del}) \in \texttt{eff}$,
            \begin{itemize}
                \item replacing \texttt{cond} by $\texttt{cond}^\prime_{\mT^\prime}$,
                \item replacing \texttt{add} by
                \begin{align*}
                    &\{\textit{ins}\_\textit{p}\_\textit{request}(\Vec{x}) \mid p(\Vec{x}) \in \texttt{add} \} \cup{} \\
                    &\{\textit{del}\_\textit{p}\_\textit{request}(\Vec{x}) \mid \lnot p(\Vec{x}) \in \texttt{del} \} \text{ and}
                \end{align*}
                \item replacing \texttt{del} by $\emptyset$.
            \end{itemize}
        \end{itemize}
        \item $\mA^\prime_{\mathsf{u}}$ contains $a_{\textit{update}}=(\texttt{pre}_{\textit{update}},\texttt{eff}_{\textit{update}})$, where:
        \begin{itemize}
            \item $\texttt{pre}_{\textit{update}} = \textit{updating}() \land \neg \textit{incompatible}\_\textit{update}()$ and
            \item for each predicate $p\in\mP$, $\texttt{eff}_{\textit{update}}$ contains
            \begin{itemize}
                \item %$e_{\textit{ins\_p}} = 
                $(\Vec{x},\ \textit{ins\_p}(\Vec{x}),\ \{p(\Vec{x})\},\ \emptyset)$,
                \item %$e_{del\_p} =
                $(\Vec{x},\ \textit{del\_p}(\Vec{x}),\ \emptyset,\ \{\lnot p(\Vec{x})\})$,
                \item %$e_{\textit{ins\_p}}' =
                $(\Vec{x},\ \textit{ins\_p\_request}(\Vec{x}),\ \emptyset,\ \{\lnot \textit{ins\_p\_request}(\Vec{x})\})$ and
                \item %$e_{\textit{del\_p}}' =
                $(\Vec{x},\ \textit{del\_p\_request}(\Vec{x}),\ \emptyset,\ \{\lnot \textit{del\_p\_request}(\Vec{x})\})$.
            \end{itemize}
        \end{itemize}
        \item $\mG^\prime_{\mathsf{u}} = \lnot \textit{updating}() \land \mG.$
    \end{enumerate}
\end{definition}
% 

% We refer to atoms of the form $\textit{ins}\_\textit{p}\_\textit{request}(\Vec{x})$ or $\textit{del}\_\textit{p}\_\textit{request}(\Vec{x})$  as \emph{request atom}.

% We note that $\mT$-consistency is ensured at each state transition by $\mR_{\mU,\mT}$ and $\textit{incompatible}\_\textit{update}()$ \cite{DBLP:journals/jair/GiacomoORS21}.

To show that this compilation scheme is correct, we first observe that the rewritings of the ECQs in conditions are not affected by the rules for the coherence update semantics.
The full proof can be found in the supplementary material.

\begin{lemma}\label{lem:entailment}
    Let $Q(\Vec{x})$ be an ECQ and $Q'_{\mT^\prime}(\Vec{x})$ be the FO-formula resulting from rewriting $Q'$ w.r.t.~$\mT'$. Then, for any assignment $\theta$ and state $s$ that is consistent with $\mT$, we have
    \[s, \mT, \theta \models Q(\Vec{x}) ~~ \text{iff} ~~ \mR^\prime_{\mathsf{u}}(s) \models Q'_{\mT^\prime}(\theta(\Vec{x}))\]
\end{lemma}

\begin{sketch}
   This follows from the fact that the additional predicates introduced in $\mR_\mT^{\mathsf{u}}$ do not interfere with the Datalog$^\neg$ rules $\mR_{\mT',Q'}\subseteq\mR'$ and the rewriting~$Q'_{\mT'}$ of $Q'$ w.r.t.~$\mT'$. 
\end{sketch}

Using Lemma \ref{lem:entailment}, we can show that the compilation scheme is correct.

\begin{theorem}\label{theo:correctness}
    Definition \ref{def:ekab_to_pddl} describes a polynomial coherence compilation scheme from ceKABs to PDDL that preserves plan size linearly.
\end{theorem}

\begin{proof}
    The Datalog$^\neg$ program $\mR'$ and the rewritings $Q'_{\mT'}$ can be constructed in polynomial time since \mT is a \DLLite TBox \cite{DBLP:conf/ijcai/CalvaneseGLLR07a,DBLP:conf/aaai/EiterOSTX12}.
    The set $\mR^{\mathsf{u}}_\mT$ can also be constructed in polynomial time \cite{DBLP:journals/jair/GiacomoORS21} and the remaining constructions in Definition~\ref{def:ekab_to_pddl} are linear in the size of the input ceKAB.

We now prove that there is a coherence plan $\Pi$ for~\mE iff there is a plan~$\Pi'$ for~$F(\mE)$.

\smallskip
\noindent $\Rightarrow:$\ \ Assume that there is a coherence plan $\Pi = \theta_1(a_1)$, $\dots$, $\theta_n(a_n)$ for $\mE$, where $\theta_i$ ($i \in \{1,\dots,n\}$) are assignments for grounding the action parameters, and let $s_0 = \mI, s_1, \dots, s_n$ be the resulting sequence of states of $\mE$, where $s_n,\mT\models\mG$. Let $\Pi^\prime$ be a PDDL plan obtained as follows: First, we replace each action $a_i$ with the corresponding action $a_i^\prime$ from Definition~\ref{def:ekab_to_pddl}(2), using the same assignment $\theta_i$. 
Second, we add the ground action $a_{\textit{update}}$ from Definition~\ref{def:ekab_to_pddl}(3) after every action $\theta_i(a_i')$.
Then, the constructed sequence of ground actions $\Pi^\prime$ is of the form $\theta_1(a_1')$, $a_{\textit{update}}$, $\dots$, $\theta_n(a^\prime_n)$, $a_{\textit{update}}$. 
Let $s^{\textit{update}}_0 = \mI, s'_1, s^{\textit{update}}_1, \dots, s'_n, s_n^{\textit{update}}$ be the sequence of states resulting from~$\Pi'$.
We show that this sequence actually exists, \ie every action is applicable in the corresponding state, and that, for every $i \in \{1, \dots, n \}$, the state $s_i^{\textit{update}}$ of $F(\mE)$ is equivalent (w.r.t.~\mT) to the state $s_{i}$ of $\mE$, via induction on the length of $\Pi$.
\[\equivto{s_0}{s^{\textit{update}}_0,}, \equivto{s_1}{s'_1,\ s^{\textit{update}}_1}, \dots, \equivto{s_n}{s'_n,\ s^{\textit{update}}_n}\]
If $\norm{\Pi} = 0$, then $s^{\textit{update}}_0 = s_0 = \mI$, and thus the claim holds.
Assuming that the claim holds for $\norm{\Pi} = n$, we now prove it for $\norm{\Pi} = n+1$.

By the induction hypothesis, $s^{\textit{update}}_n$ is equivalent (w.r.t.~\mT) to~$s_n$, which does not contain any \textit{request} predicates, and thus $\mR'_\mathsf{u}(s^{\textit{update}}_n)\not\models updating()$.
Since $\theta_{n+1}(a_{n+1})$ is applicable in $s_n$, we further know that $s_n,\mT,\theta_{n+1}\models\texttt{pre}$, where \texttt{pre} is the precondition of $a_{n+1}$.
Since $s^{\textit{update}}_n$ and~$s_n$ entail the same assertions, they also entail the same UCQs and ECQs, and thus, in particular, $s^{\textit{update}}_n,\mT,\theta_{n+1}\models\texttt{pre}$.
By Lemma~\ref{lem:entailment} and the fact that the precondition of $a'_{n+1}$ is $\lnot \textit{updating}()\land\texttt{pre}'_{\mT'}$, we obtain that $\theta_{n+1}(a'_{n+1})$ is applicable in $s_n^{\textit{update}}$.

We now consider the state $s_{n+1}'$ resulting from applying $\theta_{n+1}(a'_{n+1})$ to $s_n^{\textit{update}}$.
Due to Lemma~\ref{lem:entailment}, the request action $\theta_{n+1}(a'_{n+1})$ adds exactly the \textit{request} atoms corresponding to the update $\mU_{\theta_{n+1}(a_{n+1})}$ constructed as in Definition~\ref{def:update_from_actions}.
We can assume \wlg that this actually adds or removes at least one ground atom, and thus $\mR'_{\mathsf{u}}(s_{n+1}')\models\textit{updating}()$.
Moreover, since $\mU_{\theta_{n+1}(a_{n+1})}$ is compatible with~\mT and $\mR'_{\mathsf{u}}$ contains $\mR_\mT^{\mathsf{u}}$, by Proposition~\ref{pro:coherence_update_semantics_correctness}, $\lnot \textit{incompatible}\_\textit{update}()$ is true in $\mR'_{\mathsf{u}}(s'_{n+1})$. 
Hence, the precondition of $a_{\textit{update}}$ is satisfied, rendering $a_{\textit{update}}$ applicable in $s'_{n+1}$.

Similarly, by Proposition~\ref{pro:coherence_update_semantics_correctness}, $\mR_\mT^{\mathsf{u}}$ derives exactly the facts for inserting and deleting ground atoms $p(\Vec{c})$ (\ie $ins\_p(\Vec{c})$ or $del\_p(\Vec{c})$) that are required for computing the result of the update $\mU_{\theta_{n+1}(a_{n+1})}$ on $\langle\mT,s_{n}^{\textit{update}}\rangle$.
Since $cl_\mT(s_n)=cl_\mT(s_n^{\textit{update}})$, this result is equivalent to the result $\langle\mT,s_{n+1}\rangle$ of $\mU_{\theta_{n+1}(a_{n+1})}$ on $\langle\mT,s_{n}\rangle$.
Subsequently, the action $a_{\textit{update}}$ implements these changes on the atoms $p(\vec{c})$ and deletes all \textit{request} atoms from $s_{n+1}'$, obtaining the state $s_{n+1}^{\textit{update}}$ that is equivalent to $s_{n+1}$, as claimed.

Finally, it follows from Lemma~\ref{lem:entailment} and the above claim that the goal $\mG'_{\mathsf{u}}=\lnot \textit{updating}()\land\mG$ is satisfied by $\mR'_{\mathsf{u}}(s_n^{\textit{update}})$, i.e.\ $\Pi'$ is a plan for $F(\mE)$.
Incidentally, this shows that plan size is preserved linearly, namely by a factor of~$2$, since $\Pi'$ has exactly twice the amount of actions as~$\Pi$.

\smallskip
\noindent $\Leftarrow:$ \ \ Let $\Pi^\prime$ be a plan for $F(\mE)$. We show that $\Pi'$ must have the form $\theta_1(a'_1), a_{\textit{update}}, \dots, \theta_n(a'_n), a_{\textit{update}}$ (for some assignments $\theta_i$, $i \in \{1,\dots,n\}$) and thus correspond to a plan $\Pi = \theta_1(a_1), \dots, \theta_n(a_n)$ for $\mE$ as above.
Assuming again \wlg that each ground action $\theta_i(a'_i)$ actually adds at least one \textit{request} atom, each such action causes $\textit{updating}()$ to become true, whereas $a_{\textit{update}}$ causes $\textit{updating}()$ to become false.
Therefore, these kinds of actions must alternate.
Moreover, since $\mI$ does not contain any \textit{request} atom, and $\mG'_{\mathsf{u}}$ requires $\textit{updating}()$ to be false, the plan must begin with an action of the form $\theta_1(a_1')$ and end with $a_{\textit{update}}$.

Let $s^{\textit{update}}_0 = \mI, s'_1, s^{\textit{update}}_1, \dots, s'_n, s_n^{\textit{update}}$ and $s_0 = \mI, s_1, \dots, s_n$ be the sequences of states resulting from $\Pi'$ and $\Pi$, respectively.
As for the other direction, we show that $\Pi$ is applicable in $s_0$ and that any state $s_i$ is equivalent (w.r.t.~\mT) to the state $s_i^{\textit{update}}$, by induction on~$n$.
% 
% \[\equivto{s^{\textit{update}}_0}{s_0}, s'_1, \equivto{s^{\textit{update}}_1}{s_1}, \dots, \equivto{s^{\textit{update}}_n}{s_n}\]
% 
In the base case where $n = 0$, it is clear that both tasks share the same state $\mI = s_0 = s^{\textit{update}}_0$.

For the induction step from $n$ to $n + 1$, assume that $s_n$ is equivalent to $s_n^{\textit{update}}$ w.r.t.~\mT.
Since $\theta_{n+1}(a_{n+1}')$ is applicable in $s_n^{\textit{update}}$, we know that $\mR'_{\mathsf{u}}(s_n^{\textit{update}})\models\texttt{pre}'_{\mT'}$, where \texttt{pre} is the precondition of $a_{n+1}$.
By Lemma~\ref{lem:entailment}, we obtain $s_n^{\textit{update}},\mT,\theta_{n+1}\models\texttt{pre}$, and thus $s_n,\mT,\theta_{n+1}\models\texttt{pre}$.
Since $\mU_{\theta_{n+1}(a_{n+1})}$ is exactly characterized by the \textit{request} atoms added to $s_{n+1}'$ by $\theta_{n+1}(a_{n+1}')$ (see Definition~\ref{def:ekab_to_pddl}) and $a_{\textit{update}}$ is applicable in~$s_{n+1}'$, we know that $\mR'_{\mathsf{u}}(s_{n+1}')\not\models\textit{incompatible\_update}()$.
By Proposition~\ref{pro:coherence_update_semantics_correctness}, it follows that $\mU_{\theta_{n+1}(a_{n+1})}$ is compatible with~\mT.
This shows that $\theta_{n+1}(a_{n+1})$ is applicable in~$s_n$.

Similarly, since the subsequent $a_{\textit{update}}$ action translates the $\textit{ins\_p}(\vec{c})$ and $\textit{del\_p}(\vec{c})$ atoms in $\mR'_{\mathsf{u}}(s_{n+1}')$ into adding or deleting $p(\vec{c})$, respectively, we know that $\langle\mT,s^{\textit{update}}_{n+1}\rangle$ is equivalent to $\langle\mT,s^{\textit{update}}_n\rangle\bullet\mU_{\theta_{n+1}(a_{n+1})}$, which is equivalent to $\langle\mT,s_n\rangle\bullet\mU_{\theta_{n+1}(a_{n+1})}$ by the inductive hypothesis, and thus to $\langle\mT,s_{n+1}\rangle$ by Definition~\ref{def:update_from_actions}.
This concludes the proof of the claim.

Finally, similar to the other direction, it follows from Lemma~\ref{lem:entailment} and the fact that the final states of the two plans are equivalent that $\mG$ is satisfied in the last state~$s_n$.
\end{proof}

Extending this result to more expressive description logics is part of future work. However, we already know that, unless $\ExpTime^{\NP} = \ExpTime$, there cannot be a polynomial compilation scheme that preserves plan size polynomially for an extension of ceKABs based on Horn-$\mathcal{SROIQ}$ ontologies. The proof of the same result for eKABs \cite[Theorem~5]{DBLP:conf/aaai/Borgwardt0KKNS22} constructs a family of eKAB tasks containing a single action that adds a fresh nullary predicate $g()$ to the state. Since this predicate has no connection to the constructed ontologies, it does not matter which semantics we use for the action effects, and thus the result also holds for ceKABs.

Next, we address the \emph{coherence plan existence} problem, which decides the following:
Given a \DLLite ceKAB task~$\mE$, is there a coherence plan~$\Pi'$ for~$\mE$? For this, we use a result of \inlinecite{DBLP:journals/ai/ErolNS95} on the \emph{plan existence} problem for classical planning (PDDL without derived predicates), which the authors showed to be \ExpSpace-complete.

\begin{theorem}\label{theo:complexity}
    The coherence plan existence problem is \ExpSpace-complete. 
\end{theorem}

\begin{sketch}
    Membership holds since Definition~\ref{def:ekab_to_pddl} is a polynomial compilation scheme into PDDL with derived predicates, for which the problem in \ExpSpace. This can be shown by a small adaptation of the proof of \inlinecite{DBLP:journals/ai/ErolNS95} since evaluating derived predicates can be done in the same way as applying actions, and hence does not require more than exponential space.
    
    Hardness is shown via a polynomial reduction from the plan existence problem for PDDL. Specifically, a PDDL task $\mE'$ without derived predicates corresponds to a ceKAB $\mE$ obtained by interpreting all FO-formulas as ECQs and adding an empty TBox $\mT$. Additionally, one needs to prevent the case that action effects delete and insert the same fact, since the coherence update semantics forbids this. However, this can be avoided by adding conditional comparisons of affected facts.
\end{sketch}

% \stefan{Include new coherence update compilation? If we don't have an implementation of this, we need to argue why it is useful: integrate reasoning fully with the planner (no need for Nemo); don't add many redundant axioms via ins\_closure predicates (especially if the concept hierarchy is deep); starting point for dealing with DL-Lite$_{horn}$ ($\mT$-closure would be exponential)}

%% file: parts/exp.tex
\section{Experiments}

We conducted a range of experiments to evaluate the feasibility of our compilation and its performance compared to the pure eKAB semantics.
All code and additional benchmark data is available in the supplementary material.
% \stefan{make all data and code available}

\subsection{Implementation}
We extend the existing implementation of the eKAB compilation\footnote{\url{https://gitlab.perspicuous-computing.science/a.kovtunova/pddl-horndl}} \cite{DBLP:conf/aaai/Borgwardt0KKNS22}, which uses \emph{Clipper}\footnote{\url{https://github.com/ghxiao/clipper}} \cite{DBLP:conf/aaai/EiterOSTX12} to rewrite the UCQs in ECQs. Since they have the same syntax, eKAB and ceKAB tasks are encoded in the same way, as a combination of a \verb|.pddl| domain file, a \verb|.pddl| problem file and a \verb|.ttl| file for the ontology. Before the actual compilation, we compute the deductive closure $cl(\mT)$ using the Datalog engine \emph{Nemo}\footnote{\url{https://knowsys.github.io/nemo-doc/}} \cite{Ivliev_Nemo_Your_Friendly_2024}, based on the derivation rules of \inlinecite{DBLP:conf/otm/BorgidaCR08}. The set $cl(\mT)$ is then used to construct the update rules~$\mR_\mT^{\mathsf{u}}$. The output of the compilation is a standard PDDL task, consisting of two \verb|.pddl| files without a TBox.

%In addition to the compilation scheme from Definition~\ref{def:ekab_to_pddl} (\emph{Variant~0}), we explored three logically equivalent variants of the compilation. \emph{Variant~1} replaces the predicate \textit{incompatible\_update} with a negated form \textit{compatible\_update} to avoid the negated derived atom $\lnot\textit{incompatible\_update}()$ in the precondition of $a_{\textit{update}}$ \tagtext{var:cu}{(A)} and replaces all the rules deriving $\textit{updating}()$ by directly inserting/deleting $\textit{updating}()$ in action effects \tagtext{var:ua}{(B)}. \emph{Variants 2 and 3} utilize only \linktext{var:cu} and \linktext{var:ua}, respectively.

In addition to the compilation scheme from Definition~\ref{def:ekab_to_pddl} (variant \varzero), we explored a logically equivalent variant of the compilation. Variant \vartwo\ replaces all the rules deriving $\textit{updating}()$ by directly adding/deleting $\textit{updating}()$ in action effects.
Orthogonally to these variants, we also consider syntactically simpler conditions (in axiom bodies, action preconditions, action effects and the goal) constructed by means of a Tseitin transformation \cite{Tseitin1983}.

\subsection{Benchmarks}
% \stefan{Should we include the benchmark generator scripts in the supplementary material?}
% 
We adapted the classical Blocks benchmark and the eKAB benchmarks for DL-Lite from \inlinecite{DBLP:conf/aaai/Borgwardt0KKNS22}: Cats, Robot, Elevator, TPSA, VTA, VTA-Roles, and TaskAssign.

The Blocks instances are similar to our running example, except that there is a concept name $\ex{Holding}\sqsubseteq\ex{Blocked}$ for blocks that are currently held in the hand, and two actions \ex{pick\text{-}up} and \ex{put\text{-}down} instead of a single \ex{move} action, which is closer to the original benchmark. We used the instances of the classical Blocks domain\footnote{\url{https://github.com/aibasel/downward-benchmarks/tree/master/blocks}} to construct the problem files using a simple syntactic transformation.

Since the original instances of Cats and Robot do not have coherence plans, we used modified benchmarks that we denote by Cats$^\star$ and Robot$^\star$. In Cats$^\star$, preconditions and effects are more explicit than in Cats, ensuring that packages are consistently disarmed or removed. Actions in the original Robot benchmark take advantage of the eKAB semantics, which allow insertion and deletion of the same facts, but insertion takes precedence. Since ceKABs do not allow this, Robot$^\star$ explicitly makes the insertion and deletion effects disjoint. After these modifications, all benchmarks have plans under both eKAB and ceKAB semantics. 
% Each task instance comprises a domain, problem, and ontology file.
%The domain and problem files are encoded in an extension of the PDDL 2.0 format with support for open-world semantics and the ontology file is given in the Turtle syntax.

\subsection{Evaluation}

We used Downward Lab \cite{seipp-et-al-zenodo2017} to conduct experiments with the Fast Downward planning system \cite{DBLP:journals/jair/Helmert06} on Intel Xeon Silver 4114 processors running at 2.2 GHz with a time limit of 30 minutes and a memory limit of 3 GiB per task. The compilation time was less than 1.2\,s on all variants for the instance with the largest ontology file (Robot$^\star$).
% \duy{Compilation time above! Note: Different machine! Would that be a problem?}

For determining optimal plans, we would need an admissible heuristic, but the 
informative admissible heuristics in Fast Downward do not support derived predicates. For this reason, we focus on satisficing planning, using greedy best-first search \cite{doran-michie-rsl1966} with the FF heuristic \cite{DBLP:journals/jair/HoffmannN01}. While the original heuristic does not support derived predicates, either, Fast Downward implements two variants of supporting them. Both are based on the general idea to treat derived predicates like actions, which means that they do not have to be applied. However, the naive approach can lead to weak heuristic estimates if derived predicates are required to be false, \eg in action preconditions or the goal. For this reason, the standard implementation uses a compilation that (only for the heuristic computation) adds additional derived predicates to capture the requirement that the derived predicate is false. Since this is infeasible for cyclic dependencies, the standard heuristic variant (FF) treats negative derived predicates in such cycles as undefined. As the transformation for the remaining derived predicates can still lead to a combinatorial explosion, Fast Downward also provides a more aggressive variant (\ffapprox), where all derived predicates can be made false for free. FF will typically provide better heuristic guidance, but at a higher computational cost.

To also consider the other extreme, we include experiments with the blind heuristic that assigns~$1$ to non-goal states and~$0$ to goal states. This heuristic is extremely fast to compute, but gives no guidance beyond detecting goal states. Since it is admissible, it can also be combined with the \astar algorithm \cite{hart-et-al-ieeessc1968} to obtain a configuration that guarantees optimal plans.

\begin{table*}
\input{tables/coverage}
\caption{Number of solved instances with greedy best-first search, in the middle with the additional Tseitin transformation.}
\label{tab:coverage}
\end{table*}

\subsubsection{Comparing Heuristics}

Table \ref{tab:coverage} shows the number of solved instances (coverage) for the different configurations with greedy best-first search (GBFS). The first block shows the results for our basic implementation as in Definition~\ref{def:ekab_to_pddl}.
We observe that \ffapprox\ significantly outperforms FF, which runs out of memory on the Elevator and the VTA domain, as well as 13 of the Blocks instances. On the remaining Blocks tasks, as well as the unsolved instances of Robot$^\star$ and TaskAssign, it runs out of time.
Both can be attributed to the combinatorial explosion in the heuristic computation. If we look at the heuristic estimates for the initial states (as a proxy for heuristic guidance), we observe no significant advantage for FF on most domains but significantly better estimates on Cat$^\star$ and Elevator, where \ffapprox\ has heuristic estimate $0$, providing no guidance.
In Elevator, the advantage does not outweigh the higher computational cost, but in Cat$^\star$ the advantage translates into significantly faster solving times.

The different performance of \ffapprox\ on Cats$^\star$ and Elevator may be due to the fact that their ontologies include functional roles.
Although the Blocks domain also includes the functional role \ex{on\_block}, this may not play a large role in the planning tasks, since each action checks whether the target block is free before placing another block on top.

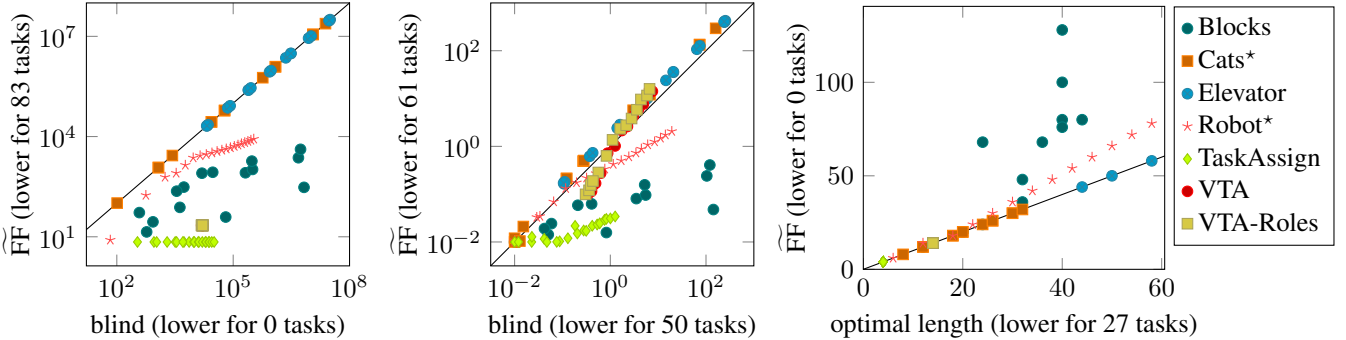
\begin{figure*}[t]
\begin{subfigure}[t]{0.285\textwidth}
\input{figures/plot_expansions}
%\caption{Expansions GBFS \ffapprox\ vs.\ blind.}
%\label{fig:expansions}
\end{subfigure}
\begin{subfigure}[t]{0.285\textwidth}
\input{figures/plot_total_time}
%\caption{Time (seconds) GBFS \ffapprox\ vs.\ blind.}
%\label{fig:time}
\end{subfigure}
\begin{subfigure}[t]{0.35\textwidth}
\input{figures/plot_cost}
%\caption{Plan cost GBFS \ffapprox\ compared to optimal plan costs.}
%\label{fig:cost}
\end{subfigure}
\caption{Analysis of \ffapprox\ (GBFS) 
on the \varzero\ variant without Tseitin transformation. Comparison to blind (GBFS) in terms of expansions (left) and time (in seconds, middle) on the jointly solved tasks; resulting plan length compared to the optimal plan length (right).}
\label{fig:scatterplots}
\end{figure*}

In the domains besides Cat$^\star$ and Elevator, \ffapprox\ indeed provides heuristic guidance, which can also be seen from the numbers of expansions in comparison to the blind heuristic, shown in Figure \ref{fig:scatterplots} (left).
The middle plot of the figure shows the resulting running times of the planner: for the Robot{$^\star$}, Blocks and TaskAssign domains, the better guidance translates into better overall performance. For VTA and VTA-Roles, the heuristic becomes more expensive to compute with increasing size of the instances, but the heuristic estimate for all initial states is $8$, so we get a relatively weaker heuristic for higher computational cost.
In terms of coverage, \ffapprox\ only has an advantage over blind search in Blocks and TPSA, but from the running time results it is to expect that, with harder tasks, it would also outperform it on Robot{$^\star$} and TaskAssign.

%FF Var0 no Tseitin
%Block (15) Robot (18) Task (11) out of time
%Block (19) Elevator (13) Trip (14) Tripv2 (9) out of memory.
\pgfplotsset{
    cycle list name=exotic
}

We also conducted experiments using \astar\ with the blind heuristic, achieving the same coverage results as with GBFS but with the guarantee that these plans are optimal. Figure \ref{fig:scatterplots} (right) plots the optimal plan length against the one with GBFS and \ffapprox\ on the commonly solved tasks. On most domains, this satisficing configuration gives optimal plans with the two exceptions of Robot$^\star$ and, most pronounced, Blocks, where the length can be more than 3 times the optimal length.
% \stefan{Should we also compare with the original Blocks benchmark (without ontology) as a baseline?}
% \gabi{I do not think that this is necessary.}
% \stefan{"cost" = "length"?}

\subsubsection{Variant for \emph{updating} Predicate}
We now turn to the impact of variant \vartwo\ shown in the second block of the table. We do not include blind search because the coverage matches variant \varzero. We observe that \vartwo\ has no impact on \ffapprox\ but leads to a significant improvement for FF: $\textit{updating}()$ is no longer a derived predicate that is required negatively in the goal and action preconditions, so it no longer causes a computational burden on the heuristic. It still is much worse than \ffapprox, in particular on the Blocks and VTA domains, but provides the best configuration so far on Elevator, solving all tasks of the domain.

\subsubsection{Tseitin Transformation}
The middle block of Table~\ref{tab:coverage} contains the analogue results with the additional Tseitin transformation. This modification leads to a slightly worse coverage with \ffapprox\ on some domains but provides an impressive boost for FF. It now solves almost all instances, and is only beaten by \ffapprox\ (without Tseitin) on Robot$^\star$ by 3 instances. Variant \vartwo\ no longer leads to an advantage in coverage, and even has a detrimental effect on \ffapprox.  Indeed, an analysis of the total solving times reveals that \ffapprox\ performs only better on Robot$^*$ and TaskAssign, but is significantly outperformed by FF on Blocks, Cat$^\star$ and Elevator.

\subsubsection{Comparing ceKABs and eKABs}
The last block of Table~\ref{tab:coverage} shows the coverage results for the analogous tasks generated with the eKAB-to-PDDL compilation by \inlinecite{DBLP:conf/aaai/Borgwardt0KKNS22} without an additional Tseitin transformation. We see that also under that compilation, \ffapprox\ performs significantly better than FF. With a Tseitin transformation, both heuristics can solve all the benchmark tasks (not included in the table), which also the best configuration could not achieve on the ceKAB-to-PDDL compilation. We conclude that supporting coherence update semantics adds extra strain to the planning system.

%% file: tables/coverage.tex
\newcolumntype{R}{>{\raggedleft\arraybackslash}X}
\begin{tabularx}{\linewidth}{lRRr|RR||RRr|RR||RR}
  % left without Tseitin transformation, middle with Tseitin, right original w/o Tseitin (ff, ffapprox)
                            & \multicolumn{3}{c|}{\varzero}
                            & \multicolumn{2}{c||}{\vartwo}
                            & \multicolumn{3}{c|}{\varzero}
                            & \multicolumn{2}{c||}{\vartwo}
                            & \multicolumn{2}{c}{eKAB} % original w/o Tseitin
                             \\[1ex]
Coverage                    & FF          & \ffapprox   & blind       & FF          & \ffapprox   & FF           & \ffapprox   & blind       & FF           & \ffapprox   & FF & \ffapprox \\
\midrule
Blocks \numtasks{34}        & 0           & \textbf{34} & 15          & 3           & \textbf{34} & \textbf{34}  & \textbf{34} & 15          & \textbf{34}  & \textbf{34} & 6  & 34                \\
Cats$^\star$ \numtasks{20}         & \textbf{20} & 14          & 14          & \textbf{20} & 14          & \textbf{20}  & 14          & 14          & \textbf{20}  & 9           & 20 & 20                    \\
Elevator \numtasks{20}      & 7           & 18          & 18          & \textbf{20} & 18          & \textbf{20}  & 18          & 18          & \textbf{20}  & 10          & 20 & 20                  \\
TPSA \numtasks{15}         & \textbf{15} & \textbf{15} & 0           & \textbf{15} & \textbf{15} & \textbf{15}  & \textbf{15} & 0           & \textbf{15}  & \textbf{15} & 15 & 15                  \\
Robot$^\star$ \numtasks{20}         & 2           & \textbf{20} & \textbf{20} & 18          & \textbf{20} & 17           & 17          & 18          & 18           & \textbf{20} & 9  & 20                      \\
TaskAssign \numtasks{20}          & 9           & \textbf{20} & \textbf{20} & 14          & \textbf{20} & \textbf{20}  & \textbf{20} & \textbf{20} & \textbf{20}  & \textbf{20} & 10 & 20                     \\
VTA \numtasks{15}          & 0           & \textbf{14} & \textbf{14} & 5           & \textbf{14} & \textbf{14}  & \textbf{14} & \textbf{14} & \textbf{14}  & \textbf{14} & 2  & 15                 \\
VTA-Roles \numtasks{15}        & 5           & \textbf{14} & \textbf{14} & 5           & \textbf{14} & \textbf{14}  & \textbf{14} & \textbf{14} & 13           & 13          & 2  & 15               \\
\midrule
\textbf{Sum \numtasks{159}} & 58          & 149         & 115         & 100         & 149         & \textbf{154} & 146         & 113         & \textbf{154} & 135         & 84 & 159
\end{tabularx}

%% file: figures/plot_expansions.tex
% expansions ffapprox vs blind with greedy search on variant 0 without Tseitin

\begin{tikzpicture}
\begin{axis}[height=2.00in, legend cell align=left, legend style={legend pos={outer north east}},
width=2.00in, xlabel={blind (lower for 0 tasks)}, xmax=100000000, xmode=log, ylabel={\ffapprox\ (lower for 83 tasks)}, ymax=100000000, ymode=log, xlabel near ticks, ylabel near ticks, ylabel shift=-5pt]
\addplot+[only marks] coordinates {
(855, 28) (381, 53) (588, 14) (4224, 76) (5344, 307) (3490, 232) (15675, 801) (64234, 39) (29697, 855) (206930, 824) (315696, 1038) (303212, 1842) (4804872, 2322) (5449600, 4151) (6703039, 302)
};
\addlegendentry{blocks}
\addplot+[only marks] coordinates {
(27599, 27599) (27591, 27591) (60050, 60050) (60074, 60074) (578120, 578120) (1230542, 1230542) (1230763, 1230763) (11414340, 11414340) (11412249, 11412249) (23959946, 23959946) (102, 102) (1180, 1180) (1183, 1183) (2706, 2706)
};
\addlegendentry{catOG}
\addplot+[only marks] coordinates {
(20789, 20789) (22576, 22576) (21128, 21128) (72197, 72197) (83522, 83522) (247875, 247875) (246060, 246060) (285837, 285837) (867849, 867849) (942205, 942205) (2271114, 2271114) (3081209, 3081209) (3071834, 3071834) (8784915, 8784915) (10195019, 10195019) (31659788, 31659788) (29533272, 29533272) (29888088, 29888088)
};
\addlegendentry{elevator}
\addplot+[only marks] coordinates {
(20297, 2915) (28344, 3249) (38501, 3590) (51086, 3928) (66458, 4279) (84973, 4653) (107048, 5092) (133082, 5574) (163528, 6099) (198840, 6629) (239521, 7193) (286053, 7789) (338997, 8432) (68, 8) (561, 175) (1754, 609) (3392, 805) (5840, 1395) (9313, 2311) (14041, 2679)
};
\addlegendentry{robot}
\addplot+[only marks] coordinates {
(5382, 7) (7313, 7) (9526, 7) (9989, 7) (12561, 7) (13081, 7) (16005, 7) (19225, 7) (19858, 7) (23437, 7) (24141, 7) (28058, 7) (32285, 7) (341, 7) (906, 7) (1065, 7) (1982, 7) (3202, 7) (3474, 7) (5046, 7)
};
\addlegendentry{task}
\addplot+[only marks] coordinates {
(16383, 22) (16384, 22) (16384, 22) (16384, 22) (16384, 22) (16384, 22) (15402, 22) (16384, 22) (16384, 22) (16182, 22) (16384, 22) (16384, 22) (16212, 22) (16363, 22)
};
\addlegendentry{trip}
\addplot+[only marks] coordinates {
(16178, 22) (16179, 22) (16179, 22) (16179, 22) (16179, 22) (16179, 22) (15201, 22) (16179, 22) (16179, 22) (15977, 22) (16179, 22) (16179, 22) (16007, 22) (16158, 22)
};
\addlegendentry{tripv2}
\draw[color=black] (axis cs:1e-70,1e-70) -- (axis cs:1e70,1e70);
\legend{}
\end{axis}
\end{tikzpicture}

%% file: figures/plot_total_time.tex
% total time ffaprox vs. blind on var0 without Tseitin

\begin{tikzpicture}
\begin{axis}[height=2.00in, legend cell align=left, legend style={legend pos={outer north east}}, 
%title=time (seconds), 
width=2.00in, xlabel={blind (lower for 50 tasks)}, xmax=1000, xmode=log, ylabel={\ffapprox\ (lower for 61 tasks)}, ymax=1000, ymode=log, xlabel near ticks, ylabel near ticks, ylabel shift=-5pt]
\addplot+[only marks] coordinates {
(0.011776, 0.01) (0.01, 0.01) (0.011158, 0.01) (0.050538, 0.014242) (0.058929, 0.024239) (0.042129, 0.019068) (0.209518, 0.058656) (0.827754, 0.015724) (0.407132, 0.062573) (3.495394, 0.080703) (5.48207, 0.095862) (5.231412, 0.15684) (103.849304, 0.240834) (120.242792, 0.407247) (142.8249, 0.047673)
};
\addlegendentry{blocks}
\addplot+[only marks] coordinates {
(0.12101, 0.212035) (0.122711, 0.213278) (0.27039, 0.491314) (0.278816, 0.504697) (3.005256, 5.690739) (6.61423, 12.654441) (6.701459, 12.033885) (73.732011, 133.524237) (73.302719, 134.679275) (159.086989, 294.398859) (0.01, 0.01) (0.01, 0.011961) (0.012875, 0.010448) (0.015224, 0.021042)
};
\addlegendentry{catOG}
\addplot+[only marks] coordinates {
(0.107558, 0.164563) (0.110775, 0.181816) (0.105616, 0.171131) (0.370588, 0.619216) (0.428517, 0.733635) (1.413674, 2.425148) (1.400594, 2.407145) (1.593497, 2.845821) (5.23225, 9.071739) (5.928271, 10.015917) (14.526911, 24.016483) (20.863129, 35.471366) (20.713815, 36.306861) (64.353185, 107.317772) (74.45434, 126.659838) (254.2241, 424.440873) (242.193628, 395.966777) (240.097348, 406.716522)
};
\addlegendentry{elevator}
\addplot+[only marks] coordinates {
(0.319015, 0.218009) (0.501815, 0.271446) (0.767509, 0.335562) (1.153952, 0.421685) (1.693009, 0.503392) (2.43583, 0.604742) (3.433248, 0.730225) (4.461057, 0.883082) (6.041628, 1.081369) (8.423206, 1.248404) (11.185739, 1.490364) (14.162066, 1.737291) (19.286312, 2.062647) (0.01, 0.01) (0.01, 0.010798) (0.029041, 0.032855) (0.033818, 0.034543) (0.06538, 0.069308) (0.116806, 0.131753) (0.196714, 0.174104)
};
\addlegendentry{robot}
\addplot+[only marks] coordinates {
(0.087337, 0.010322) (0.129876, 0.011964) (0.18371, 0.021717) (0.204649, 0.015056) (0.277583, 0.017409) (0.313138, 0.017076) (0.418785, 0.021796) (0.530805, 0.023146) (0.601125, 0.024586) (0.757338, 0.028966) (0.825149, 0.03035) (1.003308, 0.031073) (1.263722, 0.034148) (0.01, 0.01) (0.012034, 0.01) (0.022748, 0.01) (0.022968, 0.012807) (0.041553, 0.011605) (0.047269, 0.01) (0.075923, 0.010044)
};
\addlegendentry{task}
\addplot+[only marks] coordinates {
(0.629504, 0.273955) (0.879703, 0.720317) (1.258221, 1.02036) (1.740107, 2.142414) (2.279362, 2.606542) (3.010661, 4.232555) (0.344965, 0.10118) (3.905994, 6.031646) (4.917462, 7.647679) (0.404903, 0.114551) (5.734789, 11.270023) (7.573234, 14.18425) (0.430348, 0.149529) (0.492552, 0.168848)
};
\addlegendentry{trip}
\addplot+[only marks] coordinates {
(0.56249, 0.283689) (0.83367, 0.631656) (1.114753, 1.366833) (1.6123, 2.33194) (2.149156, 2.764532) (2.783763, 3.823912) (0.304667, 0.099014) (3.586184, 5.829333) (4.36405, 9.563543) (0.359403, 0.120263) (5.854922, 11.578362) (6.6109, 15.916172) (0.388663, 0.155372) (0.43317, 0.188237)
};
\addlegendentry{tripv2}
\draw[color=black] (axis cs:1e-70,1e-70) -- (axis cs:1e70,1e70);
\legend{}
\end{axis}
\end{tikzpicture}

%% file: figures/plot_cost.tex
% plan cost var0 without Tseitin

\begin{tikzpicture}[scale=1.0]
\begin{axis}[height=2.00in, width=2.2in, legend cell align=left, legend style={legend
  pos={outer north east}}, 
  xlabel={optimal length (lower for 27 tasks)}, xmax=60.6, xmin=0,
  xmode=normal, ylabel={\ffapprox\ (lower for 0 tasks)}, ymax=140.6, ymin=0, ymode=normal,
  xlabel near ticks, ylabel near ticks, ylabel shift=-5pt]
\addplot+[only marks] coordinates {
(12.0, 12.0) (20.0, 20.0) (12.0, 12.0) (24.0, 24.0) (20.0, 20.0) (32.0, 48.0) (24.0, 68.0) (20.0, 20.0) (40.0, 80.0) (40.0, 100.0) (44.0, 80.0) (40.0, 76.0) (36.0, 68.0) (40.0, 128.0) (32.0, 36.0)
};
\addlegendentry{Blocks}
\addplot+[only marks] coordinates {
(18.0, 18.0) (18.0, 18.0) (20.0, 20.0) (20.0, 20.0) (24.0, 24.0) (26.0, 26.0) (26.0, 26.0) (30.0, 30.0) (30.0, 30.0) (32.0, 32.0) (8.0, 8.0) (12.0, 12.0) (12.0, 12.0) (14.0, 14.0)
};
\addlegendentry{Cats$^\star$}
\addplot+[only marks] coordinates {
(50.0, 50.0) (58.0, 58.0) (44.0, 44.0) (68.0, 68.0) (66.0, 66.0) (78.0, 78.0) (76.0, 76.0) (80.0, 80.0) (84.0, 84.0) (86.0, 86.0) (96.0, 96.0) (90.0, 90.0) (98.0, 98.0) (102.0, 102.0) (108.0, 108.0) (110.0, 110.0) (116.0, 116.0) (114.0, 114.0)
};
\addlegendentry{Elevator}
\addplot+[only marks] coordinates {
(38.0, 48.0) (42.0, 54.0) (46.0, 60.0) (50.0, 66.0) (54.0, 72.0) (58.0, 78.0) (62.0, 84.0) (66.0, 90.0) (70.0, 96.0) (74.0, 102.0) (78.0, 108.0) (82.0, 114.0) (86.0, 120.0) (6.0, 6.0) (12.0, 14.0) (18.0, 18.0) (22.0, 24.0) (26.0, 30.0) (30.0, 36.0) (34.0, 42.0)
};
\addlegendentry{Robot$^\star$}
\addplot+[only marks] coordinates {
(4.0, 4.0) (4.0, 4.0) (4.0, 4.0) (4.0, 4.0) (4.0, 4.0) (4.0, 4.0) (4.0, 4.0) (4.0, 4.0) (4.0, 4.0) (4.0, 4.0) (4.0, 4.0) (4.0, 4.0) (4.0, 4.0) (4.0, 4.0) (4.0, 4.0) (4.0, 4.0) (4.0, 4.0) (4.0, 4.0) (4.0, 4.0) (4.0, 4.0)
};
\addlegendentry{TaskAssign}
\addplot+[only marks] coordinates {
(14.0, 14.0) (14.0, 14.0) (14.0, 14.0) (14.0, 14.0) (14.0, 14.0) (14.0, 14.0) (14.0, 14.0) (14.0, 14.0) (14.0, 14.0) (14.0, 14.0) (14.0, 14.0) (14.0, 14.0) (14.0, 14.0) (14.0, 14.0)
};
\addlegendentry{VTA}
\addplot+[only marks] coordinates {
(14.0, 14.0) (14.0, 14.0) (14.0, 14.0) (14.0, 14.0) (14.0, 14.0) (14.0, 14.0) (14.0, 14.0) (14.0, 14.0) (14.0, 14.0) (14.0, 14.0) (14.0, 14.0) (14.0, 14.0) (14.0, 14.0) (14.0, 14.0)
};
\addlegendentry{VTA-Roles}
\draw[color=black] (axis cs:-1e3,-1e3) -- (axis cs:1e3,1e3);
\end{axis}
\end{tikzpicture}

%% file: parts/concl.tex
\section{Discussion and Conclusion}

We have presented a novel combination of two semantics for planning with background ontologies: eKAB (epistemic) semantics for action conditions and coherence update semantics for single-step updates of \DLLite ABoxes, taking into account implicit effects.
We show that the resulting ceKAB semantics retains the favourable behaviour of both components, in particular allowing us to rewrite all operations into Datalog$^\neg$, and therefore into classical planning with derived predicates, in polynomial time.
This also means that the plan existence problem remains in \ExpSpace.

We conjecture that this would also hold for description logics other than $\DLLite_{core}^{(\mH\mF)}$, as long as a suitable update semantics can be defined that can be expressed as a stratified Datalog$^\neg$ program.
For example, in the Blocks domain it is natural to require that the transitive closure of \ex{on} is irreflexive \cite{grundke-et-al-icaps2024}, which could be expressed in $\DLLite_{core}^{(\mH\mF)^+}$ \cite{DBLP:journals/jair/ArtaleCKZ09}.
Other possible extensions are to allow functional roles with subroles (\eg \ex{on} should be functional), nominals (\eg \ex{Table} should be a singleton set), or (restricted) Datalog rules like $\ex{on\_block}(x,y)\leftarrow\ex{on}(x,y)\land\ex{Block}(y)$ in the ontology.
However, if the logic can express conjunction, it is unclear whether a suitable update semantics can be defined, since removing a conjunction involves a nondeterministic choice of which conjunct should be removed.
Moreover, the coherence update semantics may not always be the most appropriate choice, so we will also study combined semantics that allow to switch, \eg between eKAB and ceKAB semantics, for each insertion and deletion operation.

%Lastly, our evaluation shows that there is room for improving the compilation, for example, in the presence of functional roles (Cats$^\star$ and Elevator domains) or large ontologies (Robot).
%
Lastly, in our evaluation we have seen many instances where heuristic search does not perform better than blind search, which indicates a weak support for derived predicates in the heuristic.
This could be improved in the future by simplifying the Datalog$^\lnot$ programs used in the compilations or by developing heuristics that better support the specific structure of the resulting derived predicates.

%% file: parts/appendix.tex
\appendix

\section{Proof of Lemma~\ref{lem:entailment}}
In both directions, we use induction on the structure of~$Q$.

\noindent $\Rightarrow:$\ \ If $Q$ is of the form $[q(\vec{x})]$ for a UCQ~$q$, then similarly $Q'=[q'(\vec{x})]$.
If $s,\mT,\theta\models[q(\vec{x})]$, then $s,\mT,\theta\models q(\vec{x})$ and $\mR_{\mT,q}(s)\models P_q(\theta(\vec{x}))$ by Proposition~\ref{prop:calvanese}.
Since $\mR^\prime_{\mathsf{u}}$ contains $\mR_{\mT^\prime,q'}$ and the query predicate $P_{q'}=P_q$ as well as the rules $\{ P^\prime(\Vec{x}) \leftarrow P(\Vec{x}) \mid P \in \mP \}$, we can further infer that $\mR'_{\mathsf{u}}(s)\models P_{q'}(\theta(\vec{x})) = Q'_{\mT'}(\theta(\vec{x}))$.

In the cases where $Q$ is $p(\vec{x})$ (for a predicate $p$), $\lnot Q$, $Q_1 \land Q_2$, or $\exists y.Q$, the claim follows by the induction hypothesis and the fact that ECQs and FO-formulas share the same semantics for those constructors.

\noindent $\Leftarrow:$\ \ Conversely, for the case $Q=[q(\vec{x})]$, assume that $\mR'_{\mathsf{u}}(s)\models P_{q'}(\theta(\vec{x}))$ holds. Since the rules in $\mR_\mT^{\mathsf{u}}$ and the rules for $\textit{updating}()$ cannot infer new facts $P(\vec{c})$ or $P'(\vec{c})$ with $P\in \mP$ that may interfere with $\mR' \supseteq \mR_{\mT',q'}$, we must have $\mR'(s)\models P_{q'}(\theta(\vec{x}))$, which implies that $\mR_{\mT,q}(s)\models P_q(\theta(\vec{x}))$.
Consequently, it follows that $s, \mT, \theta \models[q(\vec{x})]$ by Proposition~\ref{prop:calvanese}.

For the other cases, we can argue similarly to the other direction, where ECQs inherit FO-semantics for the constructors $p(\vec{x})$, $\lnot$, $\land$, and $\exists$.
\qed

\section{Proof of Theorem \ref{theo:complexity}}

Membership can be shown by first constructing the polynomial-size PDDL task via the compilation scheme in Definition~\ref{def:ekab_to_pddl} and then deciding plan existence in \ExpSpace \cite{DBLP:journals/ai/ErolNS95}. 

% \textbf{Hardness:}
To show that coherence plan existence is \ExpSpace-hard, we define a polynomial reduction from the plan existence problem for PDDL without derived predicates. Let $\mE'$ be a PDDL task where the rule and derived predicate sets are empty (i.e.\ $\mR' = \mP'_{der} = \emptyset$). Then, we show that the eKAB task $\mE$ obtained by interpreting all FO-formulas in $\mE'$ as ECQs (without the bracket operator) and adding an empty TBox $\mT$ has a plan iff $\mE'$ has a plan. 
By Definition~\ref{def:coherence_plan_ekab}, $\Pi = a_1, \dots, a_n$ is a coherence plan if every associated update $\mU_{a_i}$ with a ground action $a_i$ is compatible with $\mE$ and the subsequent state $s_{i+1}$ is the result of updating $s_i$ of $\mE$ with $\mU_{a_i}$ with the coherence update semantics. 
Moreover, since $\mT = \emptyset$, Definition \ref{def:update_semantics} and Proposition \ref{pro:existence_of_abox} imply that a subsequent state $s_{i+1}$, which we obtain from the ground action $a_i$ and the state $s_i$, always accomplishes the update of $\mE$ with $\mU_i$ (which is analogous to the PDDL semantics as $\mT = \emptyset$), unless $a_i$ implies the insertion of an atom that is to be deleted in the same action. However, since $\mT = \emptyset$, we only need to deal with the case where an atom is explicitly deleted and inserted simultaneously. 

Let $a\in\mE$ be an action with effects \texttt{eff}. Then, for all $e = (\Vec{x}, \texttt{cond}, \texttt{add}, \texttt{del})$ and $e' = (\Vec{y}, \texttt{cond}', \texttt{add}', \texttt{del}')$ in \texttt{eff} where $P(\Vec{x'}) \in \texttt{add}$ and $\lnot P(\Vec{y'}) \in \texttt{del}'$ for some predicate~$P$, we substitute $e'$ with the effects
\begin{align*}
    e'_+ &= (\Vec{y}, \texttt{cond}', \texttt{add}', \texttt{del}' \setminus \{\lnot P(\Vec{y'})\}) \text{ and} \\
    e'_- &= (\Vec{y}, \texttt{cond}' \land \lnot \exists\vec{x}. (\texttt{cond} \land (\Vec{x'} = \Vec{y'})), \emptyset, \{\lnot P(\Vec{y'})\}) \,.
\end{align*}
Here, we assume that $\vec{x}$ and $\vec{y}$ are disjoint.
This can be done for all such pairs of conflicting atoms, and increases the size of the resulting effects at most polynomially, since each~$P(\vec{y'})$ can be in conflict with at most polynomially many~$P(\vec{x'})$.

We recall the semantics of PDDL and note that in the case where an action inserts and deletes the same fact, insertion takes precedence. 
Since $e$ is not modified, it is clear that the facts of the form $P(\theta(\vec{x'}))$ (for some assignment~$\theta$) are also inserted by the ceKAB task.
Moreover, the condition of $e_{-}'$ ensures that facts $P(\theta'(\vec{y'}))$ are only removed if they don't match any of the added facts $P(\theta(\vec{x'}))$, which ensures that no action inserts and deletes the same fact. Thus, by Proposition \ref{pro:existence_of_abox}, any plan of~$\mE'$ is a coherence plan of~\mE, and vice versa.
\qed

\clearpage

\section{Runtime Comparisons}

\subsection{Runtime Comparison FF vs \ffapprox}
\subsubsection{\varzero\ without Tseitin transformation}~\\
\begin{tikzpicture}
\begin{axis}[height=2.40in, legend cell align=left, legend style={legend
  pos={outer north east}}, title=total-time, width=2.40in,
  xlabel={FF}, xmax=1000, xmode=log,
  ylabel={\ffapprox}, ymax=1000, ymode=log]
\addplot+[only marks] coordinates {
(1000, 3.824355) (1000, 0.876613) (1000, 0.793335) (1000, 1.088444) (1000, 0.796377) (1000, 1.404609) (1000, 3.084584) (1000, 3.386525) (1000, 18.263274) (1000, 10.221075) (1000, 5.410307) (1000, 8.278239) (1000, 85.888933) (1000, 23.929018) (1000, 23.279927) (1000, 30.40582) (1000, 63.43987) (1000, 0.010671) (1000, 0.011258) (1000, 0.01) (1000, 0.015527) (1000, 0.025926) (1000, 0.024022) (1000, 0.067333) (1000, 0.01642) (1000, 0.069962) (1000, 0.086884) (1000, 0.107631) (1000, 0.180969) (1000, 0.274814) (1000, 0.44625) (1000, 0.050898) (1000, 0.659236) (1000, 0.195213) (1000, 0.787985)
};
\addlegendentry{Blocks}
\addplot+[only marks] coordinates {
(0.01, 0.215543) (0.01, 0.216748) (0.01, 0.494693) (0.01, 0.491399) (0.01, 5.599085) (0.01, 12.778165) (0.01, 12.4848) (0.01, 134.53181) (0.01, 134.252553) (0.01, 289.94003) (0.010438, 1000) (0.01, 1000) (0.010508, 1000) (0.010359, 1000) (0.010684, 1000) (0.012535, 1000) (0.01, 0.01) (0.01, 0.012214) (0.01, 0.012467) (0.01, 0.021952)
};
\addlegendentry{Cats$^\star$}
\addplot+[only marks] coordinates {
(0.363056, 0.165021) (0.93166, 0.180877) (1.603205, 0.174109) (5.109325, 0.627926) (11.870133, 0.736518) (25.60097, 2.381459) (46.097389, 2.360367) (1000, 2.776932) (1000, 8.977091) (1000, 10.003339) (1000, 25.242139) (1000, 35.621656) (1000, 34.270074) (1000, 108.55716) (1000, 129.843597) (1000, 423.419354) (1000, 402.415688) (1000, 416.033817) (1000, 1000) (1000, 1000)
};
\addlegendentry{Elevator}
\addplot+[only marks] coordinates {
(0.053736, 0.053834) (0.122322, 0.118179) (0.233779, 0.212349) (0.368727, 0.39339) (0.624187, 0.565129) (0.949796, 0.871764) (0.019206, 0.017476) (1.427191, 1.263179) (1.90367, 1.754318) (0.021639, 0.020963) (2.224922, 2.656231) (3.265526, 3.089709) (0.026638, 0.026866) (4.355361, 4.346491) (0.033536, 0.0327)
};
\addlegendentry{TPSA}
\addplot+[only marks] coordinates {
(1000, 0.206851) (1000, 0.270127) (1000, 0.335132) (1000, 0.414978) (1000, 0.496845) (1000, 0.604665) (1000, 0.735834) (1000, 0.835296) (1000, 0.996698) (1000, 1.250279) (1000, 1.483437) (1000, 1.740935) (1000, 2.061558) (0.457459, 0.01) (254.431571, 0.010498) (1000, 0.023938) (1000, 0.036029) (1000, 0.068545) (1000, 0.121374) (1000, 0.169825)
};
\addlegendentry{Robot$^\star$}
\addplot+[only marks] coordinates {
(80.598497, 0.012177) (451.992985, 0.013661) (1000, 0.01518) (1000, 0.01427) (1000, 0.016578) (1000, 0.019387) (1000, 0.020736) (1000, 0.022803) (1000, 0.024263) (1000, 0.027082) (1000, 0.029238) (1000, 0.031548) (1000, 0.034911) (0.01, 0.01) (0.01, 0.01) (0.0213, 0.01) (0.082635, 0.01) (0.439706, 0.01) (2.483698, 0.01) (14.009412, 0.010024)
};
\addlegendentry{TaskAssign}
\addplot+[only marks] coordinates {
(1000, 0.29821) (1000, 0.53587) (1000, 1.023526) (1000, 1.778542) (1000, 3.210735) (1000, 4.433894) (1000, 0.096433) (1000, 5.257947) (1000, 8.546427) (1000, 0.124098) (1000, 10.1721) (1000, 13.441011) (1000, 0.154267) (1000, 1000) (1000, 0.16908)
};
\addlegendentry{VTA}
\addplot+[only marks] coordinates {
(7.220972, 0.323739) (1000, 0.578325) (1000, 1.238128) (1000, 2.171351) (1000, 2.486239) (1000, 4.681526) (0.161552, 0.098868) (1000, 7.160785) (1000, 9.008187) (0.271982, 0.127301) (1000, 11.345071) (1000, 14.084685) (0.477282, 0.161425) (1000, 1000) (0.858469, 0.179836)
};
\addlegendentry{VTA-Roles}
\draw[color=black] (axis cs:1e-70,1e-70) -- (axis cs:1e70,1e70);
\end{axis}
\end{tikzpicture}

\subsubsection{\varzero\ with Tseitin transformation}~\\
\begin{tikzpicture}
\begin{axis}[height=2.40in, legend cell align=left, legend style={legend
  pos={outer north east}}, title=total-time, width=2.40in,
  xlabel={FF}, xmax=1000, xmode=log,
  ylabel={\ffapprox}, ymax=1000, ymode=log]
  \addplot+[only marks] coordinates {
(0.182372, 7.333553) (0.244304, 1.638667) (0.229915, 1.075095) (0.247797, 1.111284) (0.75801, 1.300632) (0.289729, 1.530511) (0.492493, 9.796857) (0.323907, 6.968346) (1.226464, 38.37845) (0.668601, 40.450318) (1.064734, 11.492248) (1.0847, 20.625347) (5.041116, 117.958084) (2.097469, 68.088991) (3.511621, 56.472176) (10.852529, 59.456923) (4.50061, 325.519082) (0.011201, 0.011775) (0.014858, 0.013569) (0.012079, 0.012311) (0.019431, 0.018257) (0.026524, 0.03288) (0.024682, 0.028579) (0.030404, 0.103562) (0.025951, 0.022911) (0.035791, 0.050873) (0.043566, 0.137816) (0.061789, 0.169556) (0.066256, 0.29213) (0.136747, 0.520964) (0.178011, 0.811103) (0.070656, 0.086734) (0.260838, 1.123893) (0.202773, 0.701521) (0.115851, 1.564073)
};
\addlegendentry{Blocks}
\addplot+[only marks] coordinates {
(0.01, 0.26972) (0.01, 0.273076) (0.01, 0.649991) (0.01, 0.644917) (0.01, 7.110589) (0.01, 16.422202) (0.01, 16.425408) (0.01, 172.395272) (0.01, 175.412856) (0.01, 391.918903) (0.01, 1000) (0.010514, 1000) (0.01, 1000) (0.01, 1000) (0.011092, 1000) (0.010804, 1000) (0.01, 0.01) (0.01, 0.01372) (0.01, 0.012557) (0.01, 0.025741)
};
\addlegendentry{Cats$^\star$}
\addplot+[only marks] coordinates {
(0.01, 0.204738) (0.01, 0.223298) (0.01, 0.218062) (0.011428, 0.783812) (0.01, 0.92556) (0.012562, 3.072967) (0.011598, 3.029623) (0.011319, 3.470516) (0.011185, 11.499164) (0.011994, 12.717198) (0.017927, 32.796117) (0.013821, 45.772407) (0.016044, 46.08137) (0.020188, 139.738928) (0.018184, 167.997878) (0.016699, 546.230954) (0.017399, 506.172349) (0.017464, 524.207868) (0.023293, 1000) (0.027673, 1000)
};
\addlegendentry{Elevator}
\addplot+[only marks] coordinates {
(0.054584, 0.054194) (0.113969, 0.115327) (0.222851, 0.222597) (0.386923, 0.389016) (0.603955, 0.599658) (0.998258, 0.834107) (0.017666, 0.018745) (1.300807, 1.344519) (1.945889, 1.638704) (0.022658, 0.021758) (2.329774, 2.562892) (3.070287, 3.191159) (0.027493, 0.026648) (4.265335, 4.392941) (0.033374, 0.031388)
};
\addlegendentry{TPSA}
\addplot+[only marks] coordinates {
(0.387299, 0.271236) (0.525501, 0.345833) (0.770396, 0.432362) (1.190299, 0.534271) (2.21411, 0.658153) (4.628545, 0.793207) (11.89976, 0.990638) (32.205013, 1.197489) (87.775474, 1.349329) (245.376235, 1.708496) (695.303353, 2.036566) (1000, 1000) (1000, 1000) (0.01, 0.01) (0.013532, 0.01) (0.040781, 0.025719) (0.065232, 0.040978) (0.138899, 0.081226) (0.219305, 0.157867) (0.289225, 0.214209)
};
\addlegendentry{Robot$^\star$}
\addplot+[only marks] coordinates {
(0.051648, 0.038587) (0.069232, 0.05025) (0.095782, 0.061409) (0.130639, 0.078025) (0.168684, 0.099557) (0.242782, 0.126457) (0.290297, 0.139039) (0.419012, 0.182078) (0.525575, 0.230862) (0.689263, 0.22217) (0.903243, 0.289184) (1.208317, 0.320764) (1.479779, 0.432352) (0.01, 0.01) (0.01, 0.01) (0.010653, 0.010705) (0.014486, 0.013674) (0.020327, 0.018268) (0.027603, 0.022065) (0.037317, 0.030711)
};
\addlegendentry{TaskAssign}
\addplot+[only marks] coordinates {
(0.267794, 0.266357) (0.556634, 0.707699) (1.32738, 1.112656) (2.004927, 2.134712) (2.872758, 2.512838) (4.628236, 3.972038) (0.109256, 0.096001) (5.365319, 5.684732) (8.420281, 8.704286) (0.118022, 0.128695) (10.322585, 11.346785) (14.626118, 13.648553) (0.162939, 0.141876) (1000, 1000) (0.212263, 0.166892)
};
\addlegendentry{VTA}
\addplot+[only marks] coordinates {
(0.283724, 0.350602) (0.733075, 0.705088) (1.244874, 1.247909) (2.111407, 2.307511) (2.654348, 2.84252) (4.482431, 4.107549) (0.102793, 0.102627) (6.613222, 6.445446) (8.799701, 8.777993) (0.133035, 0.119115) (11.492836, 11.757718) (14.515681, 15.397255) (0.167981, 0.15035) (1000, 1000) (0.19478, 0.173486)
};
\addlegendentry{VTA-Roles}
\draw[color=black] (axis cs:1e-70,1e-70) -- (axis cs:1e70,1e70);
\end{axis}
\end{tikzpicture}

\subsubsection{\vartwo\ without Tseitin transformation}~\\
\begin{tikzpicture}
\begin{axis}[height=2.40in, legend cell align=left, legend style={legend
  pos={outer north east}}, title=total-time, width=2.40in,
  xlabel={FF}, xmax=1000, xmode=log,
  ylabel={\ffapprox}, ymax=1000, ymode=log]
\addplot+[only marks] coordinates {
(1000, 3.99978) (1000, 0.914907) (1000, 0.821005) (1000, 1.127256) (1000, 0.826148) (1000, 1.475972) (1000, 3.179538) (1000, 3.562844) (1000, 18.036848) (1000, 9.964461) (1000, 5.635108) (1000, 9.076447) (1000, 85.96723) (1000, 25.450399) (1000, 25.078597) (1000, 32.151145) (1000, 66.608199) (1.307702, 0.010738) (1.442632, 0.01135) (1.299112, 0.01) (1000, 0.01349) (1000, 0.026806) (1000, 0.024086) (1000, 0.070555) (1000, 0.015714) (1000, 0.071848) (1000, 0.09121) (1000, 0.109104) (1000, 0.189508) (1000, 0.288823) (1000, 0.489311) (1000, 0.053666) (1000, 0.702063) (1000, 0.199661) (1000, 0.845558)
};
\addlegendentry{Blocks}
\addplot+[only marks] coordinates {
(0.01, 0.21479) (0.01, 0.214635) (0.01, 0.502339) (0.01, 0.499516) (0.01, 5.574732) (0.01, 12.680308) (0.01, 12.607024) (0.01, 131.639427) (0.01, 135.395112) (0.01, 297.010707) (0.01, 1000) (0.01, 1000) (0.01, 1000) (0.010713, 1000) (0.01239, 1000) (0.012749, 1000) (0.01, 0.01) (0.01, 0.011958) (0.01, 0.012306) (0.01, 0.028134)
};
\addlegendentry{Cats$^\star$}
\addplot+[only marks] coordinates {
(0.01034, 0.165667) (0.016938, 0.192346) (0.022574, 0.188646) (0.030746, 0.668142) (0.052772, 0.779358) (0.057732, 2.534269) (0.092594, 2.553912) (0.225123, 3.007973) (0.238525, 9.712128) (0.457582, 10.80257) (0.555174, 27.145891) (1.111089, 36.673134) (2.480886, 38.648729) (2.887974, 115.215479) (6.213088, 134.019626) (5.526367, 452.800602) (12.709614, 421.063875) (25.853477, 436.001601) (34.243803, 1000) (94.348343, 1000)
};
\addlegendentry{Elevator}
\addplot+[only marks] coordinates {
(0.05506, 0.052517) (0.114599, 0.114951) (0.212847, 0.210698) (0.35864, 0.370735) (0.578856, 0.566461) (0.950515, 0.838777) (0.01923, 0.018064) (1.361991, 1.258777) (1.850695, 1.994971) (0.022778, 0.020898) (2.349689, 2.45208) (3.068762, 3.343875) (0.02497, 0.026505) (3.989477, 3.943106) (0.034035, 0.031092)
};
\addlegendentry{TPSA}
\addplot+[only marks] coordinates {
(0.395529, 0.28092) (0.563172, 0.355501) (0.775376, 0.444637) (1.213875, 0.562647) (2.213207, 0.656974) (4.796772, 0.835409) (11.112447, 1.014356) (32.242612, 1.200348) (88.296321, 1.511763) (245.340465, 1.791949) (684.868513, 2.151821) (1000, 2.537086) (1000, 3.030444) (0.01, 0.01) (0.013029, 0.01) (0.04218, 0.025982) (0.068654, 0.044611) (0.141902, 0.08435) (0.221013, 0.161383) (0.29846, 0.222203)
};
\addlegendentry{Robot$^\star$}
\addplot+[only marks] coordinates {
(0.294729, 0.012442) (1.124748, 0.013456) (4.646556, 0.014451) (18.271759, 0.016526) (69.995665, 0.017273) (270.726395, 0.019649) (972.495885, 0.021692) (1000, 0.021243) (1000, 0.025407) (1000, 0.025625) (1000, 0.028458) (1000, 0.032994) (1000, 0.036289) (0.01, 0.01) (0.01, 0.01) (0.01, 0.01) (0.01, 0.01) (0.013537, 0.01) (0.026994, 0.01) (0.078807, 0.010648)
};
\addlegendentry{TaskAssign}
\addplot+[only marks] coordinates {
(10.496701, 0.315435) (1000, 0.738787) (1000, 1.286234) (1000, 1.645374) (1000, 2.745695) (1000, 4.323713) (0.200569, 0.098357) (1000, 6.211763) (1000, 8.587074) (0.314227, 0.111704) (1000, 10.771886) (1000, 14.539785) (0.622824, 0.154476) (1000, 1000) (1.135314, 0.165062)
};
\addlegendentry{VTA}
\addplot+[only marks] coordinates {
(7.252033, 0.288319) (1000, 0.614384) (1000, 1.431866) (1000, 2.106264) (1000, 3.201981) (1000, 4.021174) (0.159819, 0.104084) (1000, 6.457765) (1000, 8.687163) (0.257436, 0.11886) (1000, 12.299377) (1000, 15.338349) (0.443418, 0.144585) (1000, 1000) (0.826859, 0.199478)
};
\addlegendentry{VTA-Roles}
\draw[color=black] (axis cs:1e-70,1e-70) -- (axis cs:1e70,1e70);
\end{axis}
\end{tikzpicture}

\subsubsection{\vartwo\ with Tseitin transformation}~\\
\begin{tikzpicture}
\begin{axis}[height=2.40in, legend cell align=left, legend style={legend
  pos={outer north east}}, title=total-time, width=2.40in,
  xlabel={FF}, xmax=10000, xmode=log,
  ylabel={\ffapprox}, ymax=10000, ymode=log]
\addplot+[only marks] coordinates {
(0.251621, 10.765852) (0.272842, 2.42231) (0.288178, 1.981475) (0.31516, 2.950706) (0.98996, 1.955383) (0.372202, 2.274288) (0.653563, 13.443543) (0.394707, 9.759164) (0.847786, 54.403652) (0.846347, 63.653841) (1.232547, 8.546887) (1.398423, 26.825772) (5.327197, 305.919508) (2.12924, 87.267092) (3.799061, 77.943845) (14.225289, 109.382698) (5.486509, 200.661695) (0.016062, 0.016095) (0.017341, 0.018095) (0.017911, 0.016298) (0.025826, 0.02708) (0.034488, 0.044718) (0.03274, 0.043617) (0.037634, 0.144937) (0.038222, 0.031509) (0.045444, 0.075123) (0.058603, 0.299879) (0.082246, 0.215822) (0.086992, 0.438229) (0.170514, 0.693013) (0.218847, 1.361942) (0.092672, 0.122778) (0.265951, 1.776512) (0.269771, 1.055424) (0.153807, 2.286345)
};
\addlegendentry{Blocks}
\addplot+[only marks] coordinates {
(0.368698, 12.254036) (0.597357, 16.297779) (1.034532, 44.628495) (1.801955, 53.558147) (3.144796, 683.317325) (5.295424, 10000) (7.83668, 10000) (11.092919, 10000) (18.82895, 10000) (22.703475, 10000) (43.763209, 10000) (46.001067, 10000) (128.990149, 10000) (76.424466, 10000) (219.446725, 10000) (254.922349, 10000) (0.03937, 0.032302) (0.070774, 0.200285) (0.122683, 0.29865) (0.219029, 0.909608)
};
\addlegendentry{Cats$^\star$}
\addplot+[only marks] coordinates {
(0.137119, 5.842396) (0.188906, 7.404805) (0.205472, 8.205589) (0.306571, 33.820283) (0.386959, 43.382188) (0.445938, 146.958611) (0.562334, 166.003586) (0.749632, 234.484586) (1.033208, 744.318401) (1.340331, 1191.656446) (1.627869, 10000) (2.322367, 10000) (2.827381, 10000) (3.791609, 10000) (4.497741, 10000) (5.266392, 10000) (7.527563, 10000) (7.812388, 10000) (11.20352, 10000) (13.030931, 10000)
};
\addlegendentry{Elevator}
\addplot+[only marks] coordinates {
(0.12116, 0.106409) (0.230795, 0.219554) (0.440227, 0.339595) (0.685852, 0.581312) (1.076826, 0.908079) (1.551303, 1.310109) (0.04159, 0.038061) (2.385154, 1.879966) (3.129156, 2.47115) (0.049349, 0.04534) (4.03546, 3.420541) (5.581405, 4.244898) (0.058376, 0.05371) (6.731399, 5.138771) (0.070461, 0.066612)
};
\addlegendentry{TPSA}
\addplot+[only marks] coordinates {
(0.442771, 0.333052) (0.6028, 0.413004) (0.871578, 0.533025) (1.322887, 0.663081) (2.365433, 0.819424) (4.99253, 1.027924) (12.099988, 1.233037) (31.440507, 1.508952) (87.60225, 1.83661) (243.361635, 2.223384) (689.727298, 2.384732) (10000, 3.194459) (10000, 3.768822) (0.01, 0.01) (0.015393, 0.01263) (0.047068, 0.032149) (0.075991, 0.048648) (0.159691, 0.098055) (0.247216, 0.187215) (0.32967, 0.259601)
};
\addlegendentry{Robot$^\star$}
\addplot+[only marks] coordinates {
(0.061792, 0.049909) (0.080376, 0.058068) (0.107153, 0.076367) (0.148295, 0.094669) (0.193103, 0.118013) (0.236213, 0.134557) (0.302223, 0.171892) (0.431489, 0.21103) (0.524974, 0.255581) (0.774729, 0.281356) (0.836042, 0.283126) (1.090629, 0.403956) (1.601857, 0.444304) (0.01, 0.010056) (0.011204, 0.011846) (0.015604, 0.015382) (0.021602, 0.017667) (0.026952, 0.024589) (0.035587, 0.029732) (0.048657, 0.040534)
};
\addlegendentry{TaskAssign}
\addplot+[only marks] coordinates {
(0.336068, 0.282495) (0.731051, 0.613218) (1.062857, 1.096099) (2.150661, 1.972753) (2.740035, 2.96211) (4.007341, 4.210846) (0.10275, 0.101097) (5.813237, 5.453319) (8.525344, 9.07367) (0.134085, 0.124539) (11.716249, 10.776248) (14.019804, 13.925513) (0.150152, 0.150705) (10000, 10000) (0.200281, 0.17824)
};
\addlegendentry{VTA}
\addplot+[only marks] coordinates {
(0.486911, 0.393839) (0.943829, 0.94288) (1.408546, 1.363124) (2.418331, 2.152329) (3.566098, 3.834262) (5.790557, 4.886248) (0.161911, 0.156309) (8.033968, 6.995603) (11.871232, 9.817197) (0.207984, 0.177103) (12.988967, 12.555258) (10000, 10000) (0.252845, 0.203033) (10000, 10000) (0.287613, 0.262141)
};
\addlegendentry{VTA-Roles}
\draw[color=black] (axis cs:1e-70,1e-70) -- (axis cs:1e70,1e70);
\end{axis}
\end{tikzpicture}

\subsubsection{eKAB without Tseitin transformation}~\\
\begin{tikzpicture}
\begin{axis}[height=2.40in, legend cell align=left, legend style={legend
  pos={outer north east}}, title=total-time, width=2.40in,
  xlabel={FF}, xmax=10000, xmode=log,
  ylabel={\ffapprox}, ymax=10000,
  ymode=log]
\addplot+[only marks] coordinates {
(10000, 0.192302) (10000, 0.061807) (10000, 0.046612) (10000, 0.036119) (10000, 0.065336) (10000, 0.106547) (10000, 0.269466) (10000, 0.152502) (10000, 0.723868) (10000, 0.629214) (10000, 0.106737) (10000, 2.717314) (10000, 2.108034) (10000, 1.652565) (10000, 4.807289) (10000, 1.744524) (10000, 6.787013) (0.419466, 0.01) (0.432812, 0.01) (0.447355, 0.01) (435.797542, 0.01) (465.277011, 0.01) (459.830228, 0.01) (10000, 0.01) (10000, 0.01) (10000, 0.010219) (10000, 0.011864) (10000, 0.013707) (10000, 0.015683) (10000, 0.017805) (10000, 0.029252) (10000, 0.01) (10000, 0.026932) (10000, 0.030405) (10000, 0.056261)
};
\addlegendentry{Blocks}
\addplot+[only marks] coordinates {
(0.01, 0.013571) (0.01, 0.011902) (0.01, 0.022442) (0.01, 0.020159) (0.019444, 0.099792) (0.030309, 0.205247) (0.030963, 0.202363) (0.055808, 1.00877) (0.056241, 1.034361) (0.059969, 2.143044) (0.084542, 11.448049) (0.08703, 10.994923) (0.088829, 22.191542) (0.090755, 22.739371) (0.136481, 108.077712) (0.156243, 229.650001) (0.01, 0.01) (0.01, 0.01) (0.01, 0.01) (0.01, 0.01)
};
\addlegendentry{Cats$^\star$}
\addplot+[only marks] coordinates {
(0.01, 0.032892) (0.01, 0.034619) (0.01, 0.037893) (0.01, 0.104436) (0.01, 0.117141) (0.01, 0.381004) (0.01, 0.385997) (0.01, 0.419544) (0.01, 1.359599) (0.010273, 1.483188) (0.014722, 3.899338) (0.012807, 5.480189) (0.012857, 5.349643) (0.022548, 16.638454) (0.023077, 18.191017) (0.044647, 59.439303) (0.041029, 56.031124) (0.042917, 56.813237) (0.095536, 217.348784) (0.105121, 217.607831)
};
\addlegendentry{Elevator}
\addplot+[only marks] coordinates {
(0.047633, 0.046062) (0.102021, 0.097064) (0.215809, 0.1966) (0.320178, 0.322227) (0.58817, 0.598408) (0.88459, 0.916011) (0.013136, 0.015652) (1.168597, 1.208982) (1.735802, 1.553885) (0.019226, 0.018863) (2.496557, 2.225231) (3.334557, 3.185592) (0.020333, 0.023352) (3.657907, 3.668687) (0.027729, 0.026896)
};
\addlegendentry{TPSA}
\addplot+[only marks] coordinates {
(10.395168, 0.022173) (82.748867, 0.027168) (10000, 0.032068) (10000, 0.039059) (10000, 0.040307) (10000, 0.044698) (10000, 0.048877) (10000, 0.051335) (10000, 0.053566) (10000, 0.056522) (10000, 0.056106) (10000, 0.060983) (10000, 0.063323) (0.01, 0.01) (0.01, 0.01) (0.010353, 0.01) (0.024544, 0.01) (0.081658, 0.010153) (0.409365, 0.014148) (2.126467, 0.017147)
};
\addlegendentry{Robot$^\star$}
\addplot+[only marks] coordinates {
(59.01428, 0.01) (572.321533, 0.01) (1129.350865, 0.01) (10000, 0.01044) (10000, 0.010462) (10000, 0.012167) (10000, 0.013175) (10000, 0.016771) (10000, 0.01792) (10000, 0.018815) (10000, 0.020914) (10000, 0.023553) (10000, 0.026704) (0.01, 0.01) (0.01, 0.01) (0.015808, 0.01) (0.097879, 0.01) (0.214324, 0.01) (1.601633, 0.01) (6.405925, 0.01)
};
\addlegendentry{TaskAssign}
\addplot+[only marks] coordinates {
(10000, 0.222628) (10000, 0.396879) (10000, 0.770557) (10000, 1.216087) (10000, 1.982858) (10000, 3.255688) (45.034249, 0.068032) (10000, 4.476672) (10000, 6.032355) (201.672485, 0.083108) (10000, 7.066723) (10000, 10.08342) (10000, 0.10028) (10000, 12.53762) (10000, 0.134753)
};
\addlegendentry{VTA}
\addplot+[only marks] coordinates {
(10000, 0.236279) (10000, 0.490812) (10000, 0.766532) (10000, 1.193862) (10000, 2.345906) (10000, 3.367414) (45.734118, 0.068831) (10000, 4.627756) (10000, 6.124487) (200.62842, 0.085178) (10000, 7.659283) (10000, 9.385191) (10000, 0.107315) (10000, 13.094233) (10000, 0.123522)
};
\addlegendentry{VTA-Roles}
\draw[color=black] (axis cs:1e-70,1e-70) -- (axis cs:1e70,1e70);
\end{axis}
\end{tikzpicture}

\subsubsection{eKAB with Tseitin transformation}~\\
\begin{tikzpicture}
\begin{axis}[height=2.40in, legend cell align=left, legend style={legend
  pos={outer north east}}, title=total-time, width=2.40in,
  xlabel={FF}, xmode=log,
  ylabel={\ffapprox}, ymode=log]
\addplot+[only marks] coordinates {
(0.042635, 0.459602) (0.055479, 0.15249) (0.041608, 0.115732) (0.045871, 0.07081) (0.088562, 0.157523) (0.054768, 0.273264) (0.196902, 0.826414) (0.074853, 0.54126) (0.143202, 3.770684) (0.116326, 0.727485) (0.147093, 0.325272) (0.155613, 1.037011) (0.55409, 17.398732) (0.290802, 5.403999) (0.422612, 27.60474) (0.935453, 6.243649) (0.624721, 39.514638) (0.01, 0.01) (0.01, 0.01) (0.01, 0.01) (0.01, 0.01) (0.01, 0.01) (0.01, 0.01) (0.010281, 0.014025) (0.01, 0.01) (0.01, 0.016257) (0.012921, 0.01471) (0.014821, 0.020641) (0.015321, 0.023969) (0.019797, 0.032818) (0.019188, 0.05683) (0.018563, 0.01525) (0.039111, 0.088944) (0.032148, 0.042242) (0.025666, 0.145673)
};
\addlegendentry{Blocks}
\addplot+[only marks] coordinates {
(0.01, 0.015141) (0.01, 0.01671) (0.01, 0.028263) (0.01, 0.030581) (0.013316, 0.141894) (0.023387, 0.298055) (0.0238, 0.302865) (0.051907, 1.506308) (0.054976, 1.516094) (0.076743, 3.328134) (0.090375, 16.602977) (0.095702, 17.06006) (0.106881, 34.99591) (0.107179, 35.779611) (0.126994, 172.732711) (0.135586, 363.010993) (0.01, 0.01) (0.01, 0.01) (0.01, 0.01) (0.01, 0.01)
};
\addlegendentry{Cats$^\star$}
\addplot+[only marks] coordinates {
(0.01, 0.045028) (0.01, 0.047573) (0.01, 0.045075) (0.01, 0.14862) (0.01, 0.170261) (0.01, 0.548831) (0.01, 0.544825) (0.01, 0.629059) (0.01, 2.063437) (0.01, 2.236594) (0.01, 5.621749) (0.01, 8.0812) (0.01, 8.009133) (0.01, 24.465436) (0.01, 28.424508) (0.01, 93.950588) (0.01, 87.620409) (0.01, 88.965054) (0.01, 338.377011) (0.01, 344.679744)
};
\addlegendentry{Elevator}
\addplot+[only marks] coordinates {
(0.048432, 0.0485) (0.11139, 0.110952) (0.20844, 0.210479) (0.346563, 0.342284) (0.604041, 0.553239) (0.907178, 0.85725) (0.015885, 0.013753) (1.184248, 1.176737) (1.821965, 1.676167) (0.018556, 0.019209) (2.475124, 2.501071) (2.726644, 2.942869) (0.021467, 0.023123) (3.904839, 3.754094) (0.028438, 0.027984)
};
\addlegendentry{TPSA}
\addplot+[only marks] coordinates {
(0.024706, 0.023364) (0.030434, 0.026274) (0.036877, 0.033002) (0.044115, 0.038812) (0.046063, 0.043473) (0.051812, 0.045736) (0.05304, 0.052429) (0.056554, 0.054403) (0.059842, 0.056323) (0.064149, 0.059632) (0.060833, 0.06086) (0.070041, 0.062183) (0.066649, 0.068129) (0.01, 0.01) (0.01, 0.01) (0.01, 0.01) (0.01, 0.011034) (0.013115, 0.013324) (0.014682, 0.015695) (0.019746, 0.01955)
};
\addlegendentry{Robot$^\star$}
\addplot+[only marks] coordinates {
(0.045527, 0.035753) (0.062548, 0.044232) (0.084711, 0.057344) (0.116962, 0.070936) (0.163108, 0.087737) (0.22461, 0.11124) (0.29442, 0.132612) (0.353266, 0.160543) (0.484543, 0.211422) (0.698352, 0.241996) (0.819528, 0.259497) (1.171801, 0.30379) (1.380892, 0.364206) (0.01, 0.01) (0.01, 0.01) (0.01, 0.01) (0.010029, 0.01) (0.015024, 0.013423) (0.023075, 0.018938) (0.032378, 0.026202)
};
\addlegendentry{TaskAssign}
\addplot+[only marks] coordinates {
(0.223365, 0.220024) (0.399822, 0.476201) (0.865221, 0.758635) (1.509356, 1.439276) (1.911729, 2.2796) (3.195482, 3.312954) (0.069888, 0.069349) (4.034839, 4.608717) (5.847151, 6.101562) (0.086735, 0.092654) (8.033451, 7.166368) (10.824808, 10.127275) (0.101411, 0.100685) (12.875452, 12.493602) (0.121903, 0.134017)
};
\addlegendentry{VTA}
\addplot+[only marks] coordinates {
(0.229679, 0.242571) (0.48834, 0.479871) (0.846542, 0.917607) (1.260328, 1.519408) (1.749378, 2.364684) (3.343136, 2.874464) (0.076032, 0.081017) (3.987499, 4.556027) (6.270806, 5.358378) (0.087229, 0.086326) (7.967607, 8.310205) (10.549199, 10.150271) (0.11743, 0.10358) (13.392372, 13.084275) (0.121931, 0.126218)
};
\addlegendentry{VTA-Roles}
\draw[color=black] (axis cs:1e-70,1e-70) -- (axis cs:1e70,1e70);
\end{axis}
\end{tikzpicture}

\subsection{Runtime comparison \vartwo\ vs eKAB}

\subsubsection{FF without Tseitin transformation}~\\
\begin{tikzpicture}
\begin{axis}[height=2.40in, legend cell align=left, legend style={legend
  pos={outer north east}}, title=total-time, width=2.40in,
  xlabel={\vartwo}, xmax=10000, xmode=log,
  ylabel={eKAB}, ymax=10000, ymode=log]
\addplot+[only marks] coordinates {
(10000, 10000) (10000, 10000) (10000, 10000) (10000, 10000) (10000, 10000) (10000, 10000) (10000, 10000) (10000, 10000) (10000, 10000) (10000, 10000) (10000, 10000) (10000, 10000) (10000, 10000) (10000, 10000) (10000, 10000) (10000, 10000) (10000, 10000) (1.307702, 0.419466) (1.442632, 0.432812) (1.299112, 0.447355) (10000, 435.797542) (10000, 465.277011) (10000, 459.830228) (10000, 10000) (10000, 10000) (10000, 10000) (10000, 10000) (10000, 10000) (10000, 10000) (10000, 10000) (10000, 10000) (10000, 10000) (10000, 10000) (10000, 10000) (10000, 10000)
};
\addlegendentry{Blocks}
\addplot+[only marks] coordinates {
(0.01, 0.01) (0.01, 0.01) (0.01, 0.01) (0.01, 0.01) (0.01, 0.019444) (0.01, 0.030309) (0.01, 0.030963) (0.01, 0.055808) (0.01, 0.056241) (0.01, 0.059969) (0.01, 0.084542) (0.01, 0.08703) (0.01, 0.088829) (0.010713, 0.090755) (0.01239, 0.136481) (0.012749, 0.156243) (0.01, 0.01) (0.01, 0.01) (0.01, 0.01) (0.01, 0.01)
};
\addlegendentry{Cats$^\star$}
\addplot+[only marks] coordinates {
(0.01034, 0.01) (0.016938, 0.01) (0.022574, 0.01) (0.030746, 0.01) (0.052772, 0.01) (0.057732, 0.01) (0.092594, 0.01) (0.225123, 0.01) (0.238525, 0.01) (0.457582, 0.010273) (0.555174, 0.014722) (1.111089, 0.012807) (2.480886, 0.012857) (2.887974, 0.022548) (6.213088, 0.023077) (5.526367, 0.044647) (12.709614, 0.041029) (25.853477, 0.042917) (34.243803, 0.095536) (94.348343, 0.105121)
};
\addlegendentry{Elevator}
\addplot+[only marks] coordinates {
(0.05506, 0.047633) (0.114599, 0.102021) (0.212847, 0.215809) (0.35864, 0.320178) (0.578856, 0.58817) (0.950515, 0.88459) (0.01923, 0.013136) (1.361991, 1.168597) (1.850695, 1.735802) (0.022778, 0.019226) (2.349689, 2.496557) (3.068762, 3.334557) (0.02497, 0.020333) (3.989477, 3.657907) (0.034035, 0.027729)
};
\addlegendentry{TPSA}
\addplot+[only marks] coordinates {
(0.395529, 10.395168) (0.563172, 82.748867) (0.775376, 10000) (1.213875, 10000) (2.213207, 10000) (4.796772, 10000) (11.112447, 10000) (32.242612, 10000) (88.296321, 10000) (245.340465, 10000) (684.868513, 10000) (10000, 10000) (10000, 10000) (0.01, 0.01) (0.013029, 0.01) (0.04218, 0.010353) (0.068654, 0.024544) (0.141902, 0.081658) (0.221013, 0.409365) (0.29846, 2.126467)
};
\addlegendentry{Robot$^\star$}
\addplot+[only marks] coordinates {
(0.294729, 59.01428) (1.124748, 572.321533) (4.646556, 1129.350865) (18.271759, 10000) (69.995665, 10000) (270.726395, 10000) (972.495885, 10000) (10000, 10000) (10000, 10000) (10000, 10000) (10000, 10000) (10000, 10000) (10000, 10000) (0.01, 0.01) (0.01, 0.01) (0.01, 0.015808) (0.01, 0.097879) (0.013537, 0.214324) (0.026994, 1.601633) (0.078807, 6.405925)
};
\addlegendentry{TaskAssign}
\addplot+[only marks] coordinates {
(10.496701, 10000) (10000, 10000) (10000, 10000) (10000, 10000) (10000, 10000) (10000, 10000) (0.200569, 45.034249) (10000, 10000) (10000, 10000) (0.314227, 201.672485) (10000, 10000) (10000, 10000) (0.622824, 10000) (10000, 10000) (1.135314, 10000)
};
\addlegendentry{VTA}
\addplot+[only marks] coordinates {
(7.252033, 10000) (10000, 10000) (10000, 10000) (10000, 10000) (10000, 10000) (10000, 10000) (0.159819, 45.734118) (10000, 10000) (10000, 10000) (0.257436, 200.62842) (10000, 10000) (10000, 10000) (0.443418, 10000) (10000, 10000) (0.826859, 10000)
};
\addlegendentry{VTA-Roles}
\draw[color=black] (axis cs:1e-70,1e-70) -- (axis cs:1e70,1e70);
\end{axis}
\end{tikzpicture}

\subsubsection{FF with Tseitin transformation}~\\
\begin{tikzpicture}
\begin{axis}[height=2.40in, legend cell align=left, legend style={legend
  pos={outer north east}}, title=total-time, width=2.40in,
  xlabel={\vartwo}, xmax=1000, xmode=log,
  ylabel={eKAB}, ymax=1000, ymode=log]
\addplot+[only marks] coordinates {
(0.251621, 0.042635) (0.272842, 0.055479) (0.288178, 0.041608) (0.31516, 0.045871) (0.98996, 0.088562) (0.372202, 0.054768) (0.653563, 0.196902) (0.394707, 0.074853) (0.847786, 0.143202) (0.846347, 0.116326) (1.232547, 0.147093) (1.398423, 0.155613) (5.327197, 0.55409) (2.12924, 0.290802) (3.799061, 0.422612) (14.225289, 0.935453) (5.486509, 0.624721) (0.016062, 0.01) (0.017341, 0.01) (0.017911, 0.01) (0.025826, 0.01) (0.034488, 0.01) (0.03274, 0.01) (0.037634, 0.010281) (0.038222, 0.01) (0.045444, 0.01) (0.058603, 0.012921) (0.082246, 0.014821) (0.086992, 0.015321) (0.170514, 0.019797) (0.218847, 0.019188) (0.092672, 0.018563) (0.265951, 0.039111) (0.269771, 0.032148) (0.153807, 0.025666)
};
\addlegendentry{Blocks}
\addplot+[only marks] coordinates {
(0.368698, 0.01) (0.597357, 0.01) (1.034532, 0.01) (1.801955, 0.01) (3.144796, 0.013316) (5.295424, 0.023387) (7.83668, 0.0238) (11.092919, 0.051907) (18.82895, 0.054976) (22.703475, 0.076743) (43.763209, 0.090375) (46.001067, 0.095702) (128.990149, 0.106881) (76.424466, 0.107179) (219.446725, 0.126994) (254.922349, 0.135586) (0.03937, 0.01) (0.070774, 0.01) (0.122683, 0.01) (0.219029, 0.01)
};
\addlegendentry{Cats$^\star$}
\addplot+[only marks] coordinates {
(0.137119, 0.01) (0.188906, 0.01) (0.205472, 0.01) (0.306571, 0.01) (0.386959, 0.01) (0.445938, 0.01) (0.562334, 0.01) (0.749632, 0.01) (1.033208, 0.01) (1.340331, 0.01) (1.627869, 0.01) (2.322367, 0.01) (2.827381, 0.01) (3.791609, 0.01) (4.497741, 0.01) (5.266392, 0.01) (7.527563, 0.01) (7.812388, 0.01) (11.20352, 0.01) (13.030931, 0.01)
};
\addlegendentry{Elevator}
\addplot+[only marks] coordinates {
(0.12116, 0.048432) (0.230795, 0.11139) (0.440227, 0.20844) (0.685852, 0.346563) (1.076826, 0.604041) (1.551303, 0.907178) (0.04159, 0.015885) (2.385154, 1.184248) (3.129156, 1.821965) (0.049349, 0.018556) (4.03546, 2.475124) (5.581405, 2.726644) (0.058376, 0.021467) (6.731399, 3.904839) (0.070461, 0.028438)
};
\addlegendentry{TPSA}
\addplot+[only marks] coordinates {
(0.442771, 0.024706) (0.6028, 0.030434) (0.871578, 0.036877) (1.322887, 0.044115) (2.365433, 0.046063) (4.99253, 0.051812) (12.099988, 0.05304) (31.440507, 0.056554) (87.60225, 0.059842) (243.361635, 0.064149) (689.727298, 0.060833) (1000, 0.070041) (1000, 0.066649) (0.01, 0.01) (0.015393, 0.01) (0.047068, 0.01) (0.075991, 0.01) (0.159691, 0.013115) (0.247216, 0.014682) (0.32967, 0.019746)
};
\addlegendentry{Robot$^\star$}
\addplot+[only marks] coordinates {
(0.061792, 0.045527) (0.080376, 0.062548) (0.107153, 0.084711) (0.148295, 0.116962) (0.193103, 0.163108) (0.236213, 0.22461) (0.302223, 0.29442) (0.431489, 0.353266) (0.524974, 0.484543) (0.774729, 0.698352) (0.836042, 0.819528) (1.090629, 1.171801) (1.601857, 1.380892) (0.01, 0.01) (0.011204, 0.01) (0.015604, 0.01) (0.021602, 0.010029) (0.026952, 0.015024) (0.035587, 0.023075) (0.048657, 0.032378)
};
\addlegendentry{TaskAssign}
\addplot+[only marks] coordinates {
(0.336068, 0.223365) (0.731051, 0.399822) (1.062857, 0.865221) (2.150661, 1.509356) (2.740035, 1.911729) (4.007341, 3.195482) (0.10275, 0.069888) (5.813237, 4.034839) (8.525344, 5.847151) (0.134085, 0.086735) (11.716249, 8.033451) (14.019804, 10.824808) (0.150152, 0.101411) (1000, 12.875452) (0.200281, 0.121903)
};
\addlegendentry{VTA}
\addplot+[only marks] coordinates {
(0.486911, 0.229679) (0.943829, 0.48834) (1.408546, 0.846542) (2.418331, 1.260328) (3.566098, 1.749378) (5.790557, 3.343136) (0.161911, 0.076032) (8.033968, 3.987499) (11.871232, 6.270806) (0.207984, 0.087229) (12.988967, 7.967607) (1000, 10.549199) (0.252845, 0.11743) (1000, 13.392372) (0.287613, 0.121931)
};
\addlegendentry{VTA-Roles}
\draw[color=black] (axis cs:1e-70,1e-70) -- (axis cs:1e70,1e70);
\end{axis}
\end{tikzpicture}

\subsubsection{\ffapprox\ without Tseitin transformation}~\\
\begin{tikzpicture}
\begin{axis}[height=2.40in, legend cell align=left, legend style={legend
  pos={outer north east}}, title=total-time, width=2.40in,
  xlabel={\vartwo}, xmax=1000, xmode=log,
  ylabel={eKAB}, ymax=1000,
  ymode=log]
\addplot+[only marks] coordinates {
(3.99978, 0.192302) (0.914907, 0.061807) (0.821005, 0.046612) (1.127256, 0.036119) (0.826148, 0.065336) (1.475972, 0.106547) (3.179538, 0.269466) (3.562844, 0.152502) (18.036848, 0.723868) (9.964461, 0.629214) (5.635108, 0.106737) (9.076447, 2.717314) (85.96723, 2.108034) (25.450399, 1.652565) (25.078597, 4.807289) (32.151145, 1.744524) (66.608199, 6.787013) (0.010738, 0.01) (0.01135, 0.01) (0.01, 0.01) (0.01349, 0.01) (0.026806, 0.01) (0.024086, 0.01) (0.070555, 0.01) (0.015714, 0.01) (0.071848, 0.010219) (0.09121, 0.011864) (0.109104, 0.013707) (0.189508, 0.015683) (0.288823, 0.017805) (0.489311, 0.029252) (0.053666, 0.01) (0.702063, 0.026932) (0.199661, 0.030405) (0.845558, 0.056261)
};
\addlegendentry{Blocks}
\addplot+[only marks] coordinates {
(0.21479, 0.013571) (0.214635, 0.011902) (0.502339, 0.022442) (0.499516, 0.020159) (5.574732, 0.099792) (12.680308, 0.205247) (12.607024, 0.202363) (131.639427, 1.00877) (135.395112, 1.034361) (297.010707, 2.143044) (1000, 11.448049) (1000, 10.994923) (1000, 22.191542) (1000, 22.739371) (1000, 108.077712) (1000, 229.650001) (0.01, 0.01) (0.011958, 0.01) (0.012306, 0.01) (0.028134, 0.01)
};
\addlegendentry{Cats$^\star$}
\addplot+[only marks] coordinates {
(0.165667, 0.032892) (0.192346, 0.034619) (0.188646, 0.037893) (0.668142, 0.104436) (0.779358, 0.117141) (2.534269, 0.381004) (2.553912, 0.385997) (3.007973, 0.419544) (9.712128, 1.359599) (10.80257, 1.483188) (27.145891, 3.899338) (36.673134, 5.480189) (38.648729, 5.349643) (115.215479, 16.638454) (134.019626, 18.191017) (452.800602, 59.439303) (421.063875, 56.031124) (436.001601, 56.813237) (1000, 217.348784) (1000, 217.607831)
};
\addlegendentry{Elevator}
\addplot+[only marks] coordinates {
(0.052517, 0.046062) (0.114951, 0.097064) (0.210698, 0.1966) (0.370735, 0.322227) (0.566461, 0.598408) (0.838777, 0.916011) (0.018064, 0.015652) (1.258777, 1.208982) (1.994971, 1.553885) (0.020898, 0.018863) (2.45208, 2.225231) (3.343875, 3.185592) (0.026505, 0.023352) (3.943106, 3.668687) (0.031092, 0.026896)
};
\addlegendentry{TPSA}
\addplot+[only marks] coordinates {
(0.28092, 0.022173) (0.355501, 0.027168) (0.444637, 0.032068) (0.562647, 0.039059) (0.656974, 0.040307) (0.835409, 0.044698) (1.014356, 0.048877) (1.200348, 0.051335) (1.511763, 0.053566) (1.791949, 0.056522) (2.151821, 0.056106) (2.537086, 0.060983) (3.030444, 0.063323) (0.01, 0.01) (0.01, 0.01) (0.025982, 0.01) (0.044611, 0.01) (0.08435, 0.010153) (0.161383, 0.014148) (0.222203, 0.017147)
};
\addlegendentry{Robot$^\star$}
\addplot+[only marks] coordinates {
(0.012442, 0.01) (0.013456, 0.01) (0.014451, 0.01) (0.016526, 0.01044) (0.017273, 0.010462) (0.019649, 0.012167) (0.021692, 0.013175) (0.021243, 0.016771) (0.025407, 0.01792) (0.025625, 0.018815) (0.028458, 0.020914) (0.032994, 0.023553) (0.036289, 0.026704) (0.01, 0.01) (0.01, 0.01) (0.01, 0.01) (0.01, 0.01) (0.01, 0.01) (0.01, 0.01) (0.010648, 0.01)
};
\addlegendentry{TaskAssign}
\addplot+[only marks] coordinates {
(0.315435, 0.222628) (0.738787, 0.396879) (1.286234, 0.770557) (1.645374, 1.216087) (2.745695, 1.982858) (4.323713, 3.255688) (0.098357, 0.068032) (6.211763, 4.476672) (8.587074, 6.032355) (0.111704, 0.083108) (10.771886, 7.066723) (14.539785, 10.08342) (0.154476, 0.10028) (1000, 12.53762) (0.165062, 0.134753)
};
\addlegendentry{VTA}
\addplot+[only marks] coordinates {
(0.288319, 0.236279) (0.614384, 0.490812) (1.431866, 0.766532) (2.106264, 1.193862) (3.201981, 2.345906) (4.021174, 3.367414) (0.104084, 0.068831) (6.457765, 4.627756) (8.687163, 6.124487) (0.11886, 0.085178) (12.299377, 7.659283) (15.338349, 9.385191) (0.144585, 0.107315) (1000, 13.094233) (0.199478, 0.123522)
};
\addlegendentry{VTA-Roles}
\draw[color=black] (axis cs:1e-70,1e-70) -- (axis cs:1e70,1e70);
\end{axis}
\end{tikzpicture}

\subsubsection{\ffapprox\ without Tseitin transformation}~\\
\begin{tikzpicture}
\begin{axis}[height=2.40in, legend cell align=left, legend style={legend
  pos={outer north east}}, title=total-time, width=2.40in,
  xlabel={\vartwo}, xmax=10000, xmode=log,
  ylabel={eKAB}, ymax=10000,
  ymode=log]
\addplot+[only marks] coordinates {
(10.765852, 0.459602) (2.42231, 0.15249) (1.981475, 0.115732) (2.950706, 0.07081) (1.955383, 0.157523) (2.274288, 0.273264) (13.443543, 0.826414) (9.759164, 0.54126) (54.403652, 3.770684) (63.653841, 0.727485) (8.546887, 0.325272) (26.825772, 1.037011) (305.919508, 17.398732) (87.267092, 5.403999) (77.943845, 27.60474) (109.382698, 6.243649) (200.661695, 39.514638) (0.016095, 0.01) (0.018095, 0.01) (0.016298, 0.01) (0.02708, 0.01) (0.044718, 0.01) (0.043617, 0.01) (0.144937, 0.014025) (0.031509, 0.01) (0.075123, 0.016257) (0.299879, 0.01471) (0.215822, 0.020641) (0.438229, 0.023969) (0.693013, 0.032818) (1.361942, 0.05683) (0.122778, 0.01525) (1.776512, 0.088944) (1.055424, 0.042242) (2.286345, 0.145673)
};
\addlegendentry{Blocks}
\addplot+[only marks] coordinates {
(12.254036, 0.015141) (16.297779, 0.01671) (44.628495, 0.028263) (53.558147, 0.030581) (683.317325, 0.141894) (10000, 0.298055) (10000, 0.302865) (10000, 1.506308) (10000, 1.516094) (10000, 3.328134) (10000, 16.602977) (10000, 17.06006) (10000, 34.99591) (10000, 35.779611) (10000, 172.732711) (10000, 363.010993) (0.032302, 0.01) (0.200285, 0.01) (0.29865, 0.01) (0.909608, 0.01)
};
\addlegendentry{Cats$^\star$}
\addplot+[only marks] coordinates {
(5.842396, 0.045028) (7.404805, 0.047573) (8.205589, 0.045075) (33.820283, 0.14862) (43.382188, 0.170261) (146.958611, 0.548831) (166.003586, 0.544825) (234.484586, 0.629059) (744.318401, 2.063437) (1191.656446, 2.236594) (10000, 5.621749) (10000, 8.0812) (10000, 8.009133) (10000, 24.465436) (10000, 28.424508) (10000, 93.950588) (10000, 87.620409) (10000, 88.965054) (10000, 338.377011) (10000, 344.679744)
};
\addlegendentry{Elevator}
\addplot+[only marks] coordinates {
(0.106409, 0.0485) (0.219554, 0.110952) (0.339595, 0.210479) (0.581312, 0.342284) (0.908079, 0.553239) (1.310109, 0.85725) (0.038061, 0.013753) (1.879966, 1.176737) (2.47115, 1.676167) (0.04534, 0.019209) (3.420541, 2.501071) (4.244898, 2.942869) (0.05371, 0.023123) (5.138771, 3.754094) (0.066612, 0.027984)
};
\addlegendentry{TPSA}
\addplot+[only marks] coordinates {
(0.333052, 0.023364) (0.413004, 0.026274) (0.533025, 0.033002) (0.663081, 0.038812) (0.819424, 0.043473) (1.027924, 0.045736) (1.233037, 0.052429) (1.508952, 0.054403) (1.83661, 0.056323) (2.223384, 0.059632) (2.384732, 0.06086) (3.194459, 0.062183) (3.768822, 0.068129) (0.01, 0.01) (0.01263, 0.01) (0.032149, 0.01) (0.048648, 0.011034) (0.098055, 0.013324) (0.187215, 0.015695) (0.259601, 0.01955)
};
\addlegendentry{Robot$^\star$}
\addplot+[only marks] coordinates {
(0.049909, 0.035753) (0.058068, 0.044232) (0.076367, 0.057344) (0.094669, 0.070936) (0.118013, 0.087737) (0.134557, 0.11124) (0.171892, 0.132612) (0.21103, 0.160543) (0.255581, 0.211422) (0.281356, 0.241996) (0.283126, 0.259497) (0.403956, 0.30379) (0.444304, 0.364206) (0.010056, 0.01) (0.011846, 0.01) (0.015382, 0.01) (0.017667, 0.01) (0.024589, 0.013423) (0.029732, 0.018938) (0.040534, 0.026202)
};
\addlegendentry{TaskAssign}
\addplot+[only marks] coordinates {
(0.282495, 0.220024) (0.613218, 0.476201) (1.096099, 0.758635) (1.972753, 1.439276) (2.96211, 2.2796) (4.210846, 3.312954) (0.101097, 0.069349) (5.453319, 4.608717) (9.07367, 6.101562) (0.124539, 0.092654) (10.776248, 7.166368) (13.925513, 10.127275) (0.150705, 0.100685) (10000, 12.493602) (0.17824, 0.134017)
};
\addlegendentry{VTA}
\addplot+[only marks] coordinates {
(0.393839, 0.242571) (0.94288, 0.479871) (1.363124, 0.917607) (2.152329, 1.519408) (3.834262, 2.364684) (4.886248, 2.874464) (0.156309, 0.081017) (6.995603, 4.556027) (9.817197, 5.358378) (0.177103, 0.086326) (12.555258, 8.310205) (10000, 10.150271) (0.203033, 0.10358) (10000, 13.084275) (0.262141, 0.126218)
};
\addlegendentry{VTA-Roles}
\draw[color=black] (axis cs:1e-70,1e-70) -- (axis cs:1e70,1e70);
\end{axis}
\end{tikzpicture}